%% file: cl.tex
\documentclass{clv3}

\usepackage{hyperref}
\usepackage{xcolor}
\usepackage{ulem}
\definecolor{darkblue}{rgb}{0, 0, 0.5}
\hypersetup{colorlinks=true,citecolor=darkblue, linkcolor=darkblue, urlcolor=darkblue}
\usepackage{amsmath,graphicx,paralist,tabularx,multirow,color,amssymb,url,amsfonts}
\usepackage{longtable,booktabs}
\usepackage[flushleft]{threeparttable}
\usepackage{CJKutf8,setspace}

\allowdisplaybreaks

\issue{1}{1}{2018}

\runningtitle{A Joint Model of Discourse and Topics on Microblogs}

\runningauthor{Li et al.}

\begin{document}

\title{A Joint Model of Conversational Discourse and Latent Topics on Microblogs
}

\author{Jing Li\thanks{Shenzhen, China. Email: \texttt{ameliajli@tencent.com}. Jing Li is the corresponding author. This work was partially conducted when Jing Li was at Department of Systems Engineering and Engineering Management, The Chinese University of Hong Kong, HKSAR, China.}}
\affil{Tencent AI Lab}

\author{Yan Song\thanks{Shenzhen, China. Email: \texttt{clksong@tencent.com}.}}
\affil{Tencent AI Lab}

\author{Zhongyu Wei\thanks{School of Data Science, Fudan University, Shanghai, China. Email: \texttt{zywei@fudan.edu.cn}}}
\affil{Fudan University}

\author{Kam-Fai Wong\thanks{Department of Systems Engineering and Engineering Management, The Chinese University of Hong Kong, HKSAR, China. Email: \texttt{kfwong@se.cuhk.edu.hk}}}
\affil{The Chinese University of Hong Kong}

\historydates{Submission received:  October 15, 2017; revised version received: May 4, 2018; accepted for publication:  August 20, 2018.}

\maketitle

\input{sections/abstract.tex}
\input{sections/intro.tex}
\input{sections/related-work.tex}

\input{sections/model.tex}
\input{sections/exp-set.tex}
\input{sections/exp-res.tex}
\input{sections/case-study.tex}

\input{sections/summ.tex}
\input{sections/conclusion.tex}
\input{sections/appendix.tex}
\starttwocolumn
\input{sections/ack.tex}
\normalem
\bibliographystyle{compling/compling}
\bibliography{compling/compling_style}

\end{document}

%% file: sections/abstract.tex
\begin{abstract}
Conventional topic models are ineffective for topic extraction from microblog messages, because the data sparseness exhibited in short messages lacking structure and contexts results in poor message-level word co-occurrence patterns.
To address this issue, we organize microblog messages as conversation trees based on their reposting and replying relations, and propose an unsupervised model that jointly learns word distributions to represent: 1) different roles of conversational discourse, 2) various latent topics in reflecting content information.
By explicitly distinguishing the probabilities of messages with varying discourse roles in containing topical words, our model is able to discover clusters of discourse words that are indicative of topical content. 
In an automatic evaluation on large-scale microblog corpora, our joint model yields topics with better coherence scores than competitive topic models from previous studies.
Qualitative analysis on model outputs indicates that our model induces meaningful representations for both discourse and topics.
We further present an empirical study on microblog summarization based on the outputs of our joint model. 
The results show that the jointly modeled discourse and topic representations can effectively indicate summary-worthy content in microblog conversations. 
 \end{abstract}

%% file: sections/intro.tex
\section{Introduction}

Over the past two decades, the Internet has been revolutionizing the way we communicate.
Microblog, a social networking channel over the Internet, further accelerates communication and information exchange.
Popular microblog platforms, such as Twitter\footnote{\url{twitter.com}} and Sina Weibo\footnote{\url{weibo.com}}, have become important outlets for individuals to share information and voice opinions, which further benefit downstream applications such as instant detection of breaking events \cite{DBLP:conf/kdd/LinZMH10,DBLP:conf/icwsm/WengL11,DBLP:conf/cicling/PengLCHXW15}, real-time and ad-hoc search of microblog messages \cite{DBLP:conf/coling/DuanJQZS10,DBLP:conf/nlpcc/LiWWZCW15}, public opinions and user behaviors understanding on societal issues \citep{DBLP:conf/lrec/PakP10,DBLP:conf/icwsm/KouloumpisWM11,DBLP:conf/cikm/PopescuP10}, etc.

However, the explosive growth of microblog data far outpaces human beings' speed of reading and understanding.
As a consequence, there is a pressing need for effective natural language processing (NLP) systems that can automatically identify gist information, and make sense of the unmanageable amount of user-generated social media content~\cite{DBLP:series/synthesis/2015Farzindar}.
As one of the important and fundamental text analytic approaches, topic models extract key components embedded in microblog content by clustering words that describe similar semantic meanings to form latent ``topics''.
The derived intermediate topic representations have proven beneficial to many NLP applications for social media, such as summarization~\cite{DBLP:conf/icwsm/HarabagiuH11}, classification~\cite{DBLP:conf/www/PhanNH08,DBLP:conf/emnlp/ZengLSGLK2018}, and recommendation on microblogs~\cite{DBLP:conf/naacl/ZengLWBSW18}.

Conventionally, probabilistic topic models, e.g., probabilistic latent semantic analysis (pLSA) \cite{DBLP:conf/sigir/Hofmann99} and latent Dirichlet allocation (LDA) \cite{DBLP:conf/nips/BleiGJT03}, have achieved huge success over the past decade owing to their fully unsupervised manner and ease of extension. The semantic structure discovered by these topic models have facilitated the progress of many research fields, e.g., information retrieval~\cite{DBLP:journals/ftir/Boyd-GraberHM17}, data mining~\cite{DBLP:journals/widm/LinIWG15}, and natural language processing~\cite{DBLP:conf/naacl/NewmanLGB10}.
Nevertheless, ascribing to their reliance on document-level word co-occurrence patterns, the progress is still limited to formal conventional documents such as news reports \cite{DBLP:journals/jmlr/BleiNJ03} and scientific articles~\citep{DBLP:conf/uai/Rosen-ZviGSS04}. The aforementioned models work poorly when directly applied to short and colloquial texts, e.g., microblog posts, owing to severe sparsity exhibited in such text genre~\cite{DBLP:conf/kdd/WangM06,hong2010empirical}.

Previous research has proposed several methods to deal with the sparsity issue in short texts.
One common approach is to aggregate short messages into long pseudo-documents.
Many studies heuristically aggregate messages based on authorship~\cite{DBLP:conf/ecir/ZhaoJWHLYL11,hong2010empirical}, shared words~\cite{DBLP:conf/wsdm/WengLJH10}, or hashtags~\cite{DBLP:conf/icwsm/RamageDL10,DBLP:conf/sigir/MehrotraSBX13}.
\citet{DBLP:conf/ijcai/QuanKGP15} proposes a self-aggregation-based topic model that aggregate texts jointly with topic inference. 
Another popular solution is to take into account word relations to alleviate document-level word sparseness. Biterm topic model (BTM) directly models the generation of word-pair co-occurrence patterns in each individual message~\cite{DBLP:conf/www/YanGLC13,DBLP:journals/tkde/ChengYLG14}.
More recently, word embeddings trained by large-scale external data are leveraged to capture word relations and improve topic models on short texts~\cite{DBLP:conf/acl/DasZD15,DBLP:journals/tacl/NguyenBDJ15,DBLP:conf/sigir/LiWZSM16,DBLP:journals/tois/LiDWZSM17,DBLP:conf/ijcai/XunLZGZ17,DBLP:conf/sigir/ShiLJSL17}.

To date, most efforts focus on content in messages, but ignore the rich discourse structure embedded in ubiquitous user interactions on microblog platforms. On microblogs, which are originally built for user communication and interaction, conversations are freely formed on issues of interests by reposting messages and replying to others.
When joining a conversation, users generally post topically related content, which naturally provide effective contextual information for topic discovery. \citet{DBLP:conf/icwsm/Alvarez-MelisS16} has shown that simply aggregating messages based on conversations can significantly boost the performance of conventional topic models, and outperform models exploiting hashtag-based and user-based aggregations.

Another important issue ignored in most previous studies is the effective separation of topical words from non-topic ones~\cite{DBLP:conf/acl/LiLGHW16}.
In microblog content, owing to its colloquial nature, non-topic words such as sentimental (e.g., ``great'' and ``ToT''), functional (e.g., ``doubt'' and ``why''), and other non-topic words (e.g., ``oh'' and ``oops'') are common and usually mixed with topical words. The occurrence of non-topic words may distract the models from recognizing topical content, which will thus lead to the failure to produce coherent and meaningful topics. In this article, we propose a novel model that examines the entire context of a conversation and jointly explores word distributions representing varying types of topical content and \textit{discourse roles} such as agreement, question-asking, argument, and other dialogue acts~\citep{DBLP:conf/naacl/RitterCD10}.\footnote{In this work, a discourse role refers to a certain type of dialogue act on message level, e.g., agreement or argument. 
The discourse structure of a conversation means some combination (or a probability distribution) of discourse roles.} 
Though~\citet{DBLP:conf/naacl/RitterCD10} separates discourse, topic, and other words for modeling conversation content, their model focuses on dialogue act modeling and only yields one distribution for topical content. Therefore, their model is unable to distinguish varying latent topics reflecting message content underlying the corpus. 
\citet{DBLP:conf/acl/LiLGHW16} leverages conversational discourse structure to detect topical words from microblog posts, which explicitly explores the probabilities of different discourse roles that contain topical words.
However, \citet{DBLP:conf/acl/LiLGHW16} depends on a pre-trained discourse tagger and acquires a time-consuming and expensive manual annotation process for annotating conversational discourse roles on microblog messages, which does not scale for large datasets 
\cite{DBLP:conf/naacl/RitterCD10,DBLP:conf/ijcai/JotyCL11}.

To exploit discourse structure of microblog conversations, we link microblog posts using reposting and replying relations to build conversation trees. Particularly, the root of a conversation tree refers to the original post and its edges represent the reposting or replying relations. To illustrate the interplay between topic and discourse, Figure~\ref{fig:tree} displays a snippet of Twitter conversation about ``Trump administration's immigration ban". From the conversation, we can observe two major components: 1) \textit{discourse}, indicated by the underlined words, describes the intention and pragmatic roles of messages in conversation structure, such as making a statement or asking a question; 2) \textit{topic}, represented by the bold words, captures the topic and focus of the conversation, such as ``racialism'' and ``Muslims''. 
As we can see, different discourse roles vary in probabilities to contain key content reflecting the conversation focus. For example, in Figure~\ref{fig:tree}, [R5] doubts the assertion of ``immigration ban is good'', and raises a new focus on ``racialism''. This in fact contains more topic-related words than [R6], which simply reacts to its parent. For this reason, in this article, we attempt to identify messages with ``good'' discourse roles that tend to describe key focuses and salient topics of a microblog conversation tree, which enables the discovery of ``good'' words reflecting coherent topics. Importantly, our joint model of conversational discourse and latent topics is fully unsupervised, which therefore does not require any manual annotation.

\input{figures/intro-case}

For evaluation, we conduct quantitative and qualitative analysis on large-scale Twitter and Sina Weibo corpora. Experimental results show that topics induced by our model are more coherent than existing models. Qualitative analysis on discourse further shows that our model can yield meaningful clusters of words related to manually crafted discourse categories.
In addition, we present an empirical study on downstream application of microblog conversation summarization. Empirical results on ROUGE~\cite{lin2004rouge} show that summaries produced based on our joint model contain more salient information than state-of-the-art summarization systems. Human evaluation also indicates that our output summaries are competitive to existing unsupervised summarization systems in the aspects of informativeness, conciseness, and readability. 

In summary, our contributions in this article are three-fold:
\begin{itemize}
\item\textbf{Microblog posts organized as conversation trees for topic modeling.}
We propose a novel concept of representing microblog posts as conversation trees by connecting microblog posts based on \textit{reposting} and \textit{replying} relations for topic modeling. Conversation tree structure helps enrich context, alleviate data sparseness, and in turn improve topic modeling.
\item\textbf{Exploiting discourse in conversations for topic modeling.} Our model differentiates the generative process of topical and non-topic words, according to the discourse role of the message where a word is drawn from being. This helps the model in identifying the topic-specific information from non-topic background.
\item\textbf{Thorough empirical study on the inferred topic representations.} Our model shows better results than competitive topic models when evaluated on large-scale real-world microblog corpora. We also present an effective method for using our induced results on microblog conversation summarization.
\end{itemize}

%% file: figures/intro-case.tex
    \begin{figure}[ht]
   	\centering
     \includegraphics[width=9cm]{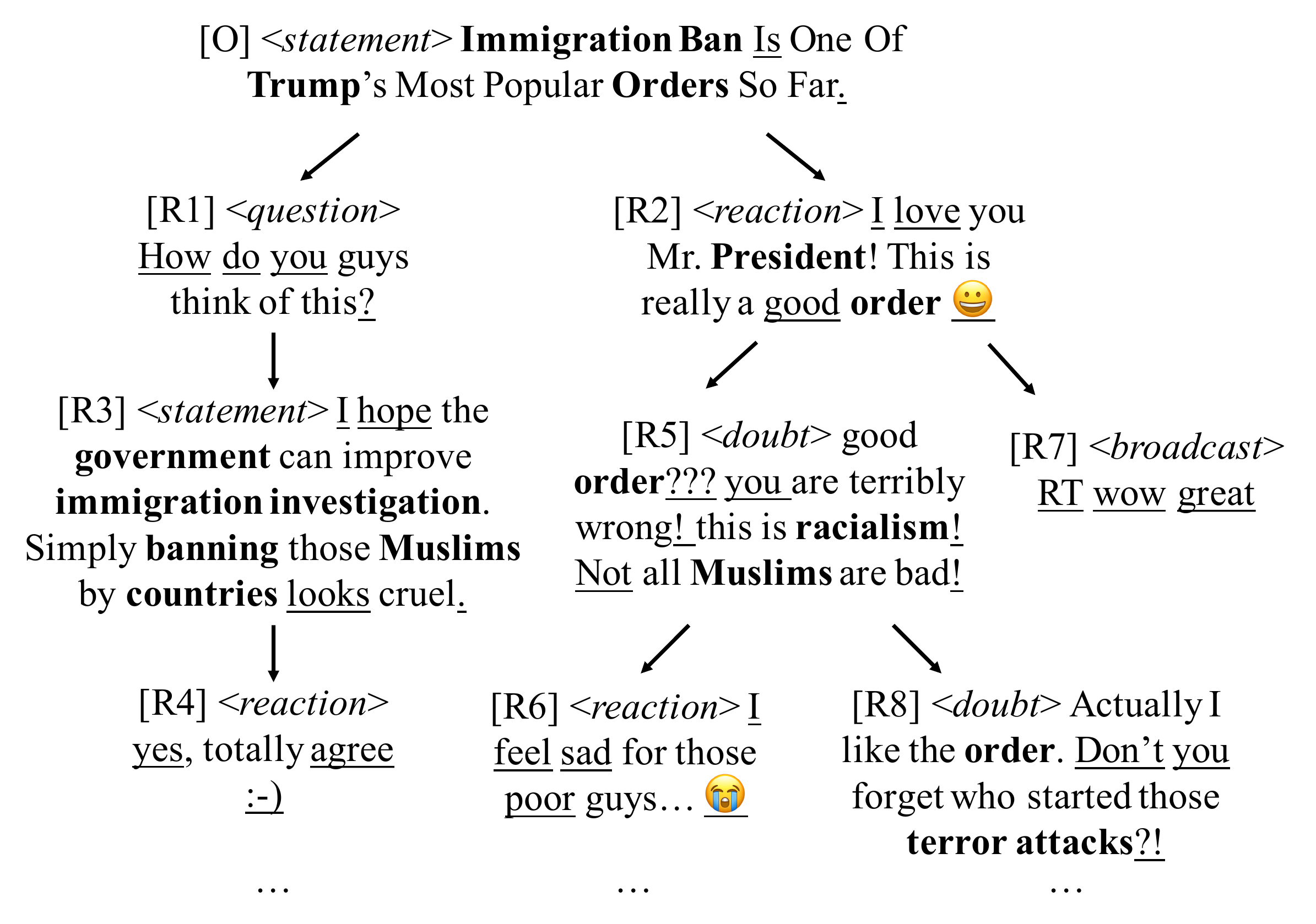}
    \caption{A sample Twitter conversation tree on ``Trump administration's immigration ban". [O]: the original post; [Ri]: the i-th repost or reply; arrow lines: reposting or replying relations; \textit{italic words} in $\langle\rangle$: discourse role of the message; \underline{underlined words}: words indicating discourse roles; \textbf{bold words}: topical words representing the discussion focus.
     }
     \vskip -0.5em
	\label{fig:tree}
    \end{figure}

%% file: sections/related-work.tex
\section{Related Work}\label{sec:related-work}

This article builds upon diverse streams of previous work in lines of \textit{topic modeling}, \textit{discourse analysis}, and \textit{microblog summarization}, 
which are briefly surveyed as follows.

\subsection{Topic Models}

Topic models aim to discover the latent semantic information, i.e., \textit{topics}, from texts and have been extensively studied.
This work is built upon the success of latent Dirichlet allocation (LDA) model \cite{DBLP:conf/nips/BleiGJT03,DBLP:journals/jmlr/BleiNJ03} and aims to learn topics in microblog messages.
We first briefly introduce LDA in Section~\ref{ssec:related:lda} and then review the related work on topic modeling for microblog content in Section~\ref{ssec:related:microblog-topic}.

\subsubsection{LDA: Springboard of Topic Models}\label{ssec:related:lda}

Latent Dirichlet allocation \cite{DBLP:journals/jmlr/BleiNJ03} is one of the most popular and well-known topic models.
It employs Dirichlet priors to generate document-topic and topic-word distributions, and has shown effective in extracting topics from conventional documents.
LDA plays an important role in semantic representation learning and serves as the springboard of many famous topics models, e.g., hierarchical latent Dirichlet allocation (HLDA)~\cite{DBLP:conf/nips/BleiGJT03}, author-topic model~\cite{DBLP:conf/uai/Rosen-ZviGSS04}, etc.
In addition to ``topic'' modeling, it has also inspired \textit{discourse} \cite{DBLP:conf/sigdial/CrookGP09,DBLP:conf/naacl/RitterCD10,DBLP:conf/ijcai/JotyCL11} detection without supervision or with weak supervision. 
However, none of the aforementioned work jointly infers discourse and topics on microblog conversations, which is a gap this article fills in.
Also, to the best of our knowledge, our work serves as the first attempt to exploit the joint effects of discourse and topic on unsupervised microblog conversation summarization.

\subsubsection{Topic Models for Microblog Posts}\label{ssec:related:microblog-topic}

Previous research has demonstrated that standard topic models, essentially focusing on document-level word co-occurrences, are not suitable for short and informal microblog messages due to the severe data sparsity exhibited in short texts~\cite{DBLP:conf/kdd/WangM06,hong2010empirical}.
As a result, one line of previous work focuses on enriching and exploiting contextual information. 
\citet{DBLP:conf/wsdm/WengLJH10}, \citet{hong2010empirical}, and \citet{DBLP:conf/ecir/ZhaoJWHLYL11} first heuristically aggregate messages posted by the same user or sharing the same words before conventional topic models are applied.
Their simple strategies, however, pose some problems.
For example, it is common that a user has various interests and thus posts messages covering a wide range of topics.
\citet{DBLP:conf/icwsm/RamageDL10} and \citet{DBLP:conf/sigir/MehrotraSBX13} employ hashtags as labels to train supervised topic models. Nevertheless, these models depend on large-scale hashtag-labeled data for model training. Moreover, their performance can be inevitably compromised when facing unseen topics that are irrelevant to any hashtag in training data. Such phenomenon are common due to the rapid change and wide variety of topics on social media.
Biterm topic model (BTM)~\cite{DBLP:conf/www/YanGLC13,DBLP:journals/tkde/ChengYLG14} directly explores unordered word-pair co-occurrence patterns in each individual message, which is equivalent to extending short documents into a biterm set consisting of all combinations of any two distinct words appearing in the document.
SATM~\cite{DBLP:conf/ijcai/QuanKGP15} combines short texts aggregation and topic induction into a unified model. However, in SATM, no prior knowledge is given to ensure the quality of text aggregation, which will further affect the performance of topic inference.

Different from the aforementioned work, we organize microblog messages as conversation trees based on reposting and replying relations. It allows us to explore word co-occurrence patterns in richer context, as messages in one conversation generally focus on relevant topics. 
Even though researchers have started to take the contexts provided by conversations into account when discovering topics on microblogs \cite{DBLP:conf/icwsm/Alvarez-MelisS16,DBLP:conf/acl/LiLGHW16}, there is much less work that jointly predicts the topical words along with the discourse structure in conversations. 
\citet{DBLP:conf/naacl/RitterCD10} models dialogue acts in conversations via separating discourse words from topical words and others. While their model produces only one word distribution to represent the topical content, our model is capable of generating varying discourse and topic word distributions. Another main difference is that our model explicitly explores the probabilities of messages with different discourse roles in containing topical words for topic representation, while their model generates topical words from a conversation-specific distribution
over word types regardless of the different discourse roles of messages.
\citet{DBLP:conf/acl/LiLGHW16} serves as another prior effort to leverage conversation structure, captured by a supervised discourse tagger, on topic induction.
Different from them, our model learns discourse structure for conversations in a fully unsupervised manner, which does not require annotated data.

Another line of research tackles the data sparseness by modeling word relations instead of word occurrences in documents. For example, recent research work has shown that distributional similarities of words captured by word embeddings \cite{DBLP:conf/nips/MikolovSCCD13,DBLP:conf/naacl/MikolovYZ13} are useful in recognizing interpretable topic word clusters from short texts~\cite{DBLP:conf/acl/DasZD15,DBLP:journals/tacl/NguyenBDJ15,DBLP:conf/sigir/LiWZSM16,DBLP:journals/tois/LiDWZSM17,DBLP:conf/ijcai/XunLZGZ17,DBLP:conf/sigir/ShiLJSL17}.
These topic models heavily rely on meaningful word embeddings needed to be trained on a large-scale high-quality external corpus, which should be both in the same domain and the same language as the data for topic modeling~\cite{DBLP:conf/acl/BollegalaMK15}. However, such external resource is not always available. For example, to the best of our knowledge, there exists no high-quality word embedding corpus for Chinese social media so far.
In contrast to these prior methods, our model does not have the prerequisite to external resource, whose general applicability in cold-start scenarios is therefore ensured.

\subsubsection{Topic Modeling and Summarization}\label{ssec:back:tm&summ}

Previous studies have shown that the topic representation captured by topic models is useful for summarization \cite{DBLP:books/sp/mining2012/NenkovaM12}.
Specifically, there are two different purposes of using topic models in existing summarization systems:
(1) to separate summary-worthy content and non-content background (general information) \cite{DBLP:conf/acl/DaumeM06,DBLP:conf/naacl/HaghighiV09,DBLP:conf/acl/CelikyilmazH10},
and (2) to cluster sentences or documents into topics, and summaries are then generated from each topic cluster for minimizing content redundancy~\cite{DBLP:journals/ipm/SaltonSMB97,DBLP:conf/aaai/McKeownKHBE99,DBLP:conf/coling/SiddharthanNM04}.
Similar techniques have also been applied to summarize events or opinions on microblogs \cite{DBLP:conf/icwsm/ChakrabartiP11,DBLP:conf/coling/DuanCWZS12,DBLP:conf/naacl/ShenLWL13,DBLP:conf/waim/LongWCJY11,rosa2011topical,DBLP:conf/kdd/MengWLZLW12}.

Our downstream application on microblog summarization lies in the research line of (1), whereas we integrate the effects of discourse on key content identification, which has not been studied in any prior work.
Also it is worth noting that, following (2) to cluster messages before summarization is beyond the scope of this work because we are focusing on summarizing a single conversation tree, on which there are limited topics. And we leave the potential of using our model to segment topics for multi-conversation summarization to future work.

\subsection{Discourse Analysis}\label{sec:back:discourse}

Discourse reflects the architecture of textual structure, where the semantic or pragmatic relations among text units (e.g., clauses, sentences, paragraphs) are defined. Here we review prior work on single document discourse analysis in Section~\ref{ssec:back:tra-discourse}, followed by a description on discourse extension to represent conversation structures in Section \ref{ssec:back:conv-discourse}. 

\subsubsection{Traditional View of Discourse}\label{ssec:back:tra-discourse}
It has been long pointed out that a coherent document, which gives readers continuity of senses \cite{de1981textlinguistics}, is not simply a collection of independent sentences.
Linguists have striven to the study of discourse analysis ever since ancient Greece \cite{bakker2009discourse}.
Early work shapes the modern concept of discourse \cite{hovy1997parsimonious} via depicting connections between text units, which reveals the structural art behind a coherent documents.

Rhetorical structure theory (RST) \cite{william1988rhetorical} is one of the most influential discourse theories. According to its assumption, a coherent document can be represented by text units at different levels (e.g., clauses, sentences, paragraphs) in a hierarchical tree structure.
In particular, the minimal units in RST, i.e., leaves of the tree structure, are defined as sub-sentential clauses, namely, elementary discourse units (EDUs). Adjacent units are linked by rhetorical relations, e.g., condition, comparison, elaboration, etc.
Based on RST, early work employs hand-coded rules for automatic discourse analysis \cite{DBLP:journals/coling/Marcu00,DBLP:conf/coling/ThanhAH04}.
Later, thanks to the development of large-scale discourse corpus, e.g., RST corpus \cite{DBLP:conf/sigdial/CarlsonMO01}, Graph Bank corpus \cite{DBLP:journals/coling/WolfG05}, and Penn Discourse Treebank (PDTB) \cite{DBLP:conf/lrec/PrasadDLMRJW08}, data-driven and learning-based discourse parsers that exploit various manually designed features~\cite{DBLP:conf/naacl/SoricutM03,DBLP:conf/conll/BaldridgeL05,DBLP:conf/naacl/SubbaE09,DBLP:conf/emnlp/LinKN09,DBLP:conf/acl/FengH14,DBLP:conf/emnlp/JotyCN12,DBLP:conf/acl/FisherR07} and representative learning \cite{DBLP:conf/acl/JiE14,DBLP:conf/emnlp/LiLH14} become popular.

\subsubsection{Discourse Analysis on Conversations}\label{ssec:back:conv-discourse}

\citet{DBLP:journals/corr/cs-CL-0006023} is one of the first studies focusing on this problem, which provides a general schema of understanding conversations with discourse analysis.
Due to the complex structure and informal language style, discourse parsing on conversations is still a challenging problem~\cite{DBLP:conf/naacl/PerretAAM16}. Most research focuses on the detection of dialogue acts (DA)\footnote{Dialogue act can be used interchangeably with speech act~\cite{DBLP:journals/corr/cs-CL-0006023}.
}, which is defined in \citet{DBLP:journals/corr/cs-CL-0006023} as the first level conversational discourse structure.
It is worth noting that, a DA represents the shallow discourse role that captures illocutionary meanings of an utterance, e.g., ``statement'', ``question'', ``agreement'', etc. 

Automatic dialogue act taggers have been conventionally trained in a supervised way with pre-defined tag inventories and annotated data \cite{DBLP:journals/corr/cs-CL-0006023,DBLP:conf/emnlp/CohenCM04,DBLP:conf/acl/BangaloreFS06}.
However, DA definition is generally domain-specific and usually involves the manual designs from experts. Also, the data annotation process is slow and expensive resulting in the limitation of data available for training DA 
classifiers~\cite{jurafsky1997switchboard,dhillon2004meeting,DBLP:conf/naacl/RitterCD10,DBLP:conf/ijcai/JotyCL11}.
These issues are pressing with the arrival of the Internet era where new domains of conversations and even new types of dialogue act tags are boomed \cite{DBLP:conf/naacl/RitterCD10,DBLP:conf/ijcai/JotyCL11}.

For this reason, researchers have proposed unsupervised or weakly-supervised dialogue act taggers that identify indicative discourse word clusters based on probabilistic graphical models \cite{DBLP:conf/sigdial/CrookGP09,DBLP:conf/naacl/RitterCD10,DBLP:conf/ijcai/JotyCL11}.
In our work, the discourse detection module is inspired by these previous models, where discourse roles are represented by word distributions and recognized in an unsupervised manner.
Different from the previous work that focuses on discourse analysis, we explore the effects of discourse structure of conversations on distinguishing varying latent topics underlying the given collection, which has not been studied before.
In addition, most existing unsupervised approaches for conversation modeling follows hidden Markov model (HMM) convention and induces discourse representations in conversation threads.
Considering that most social media conversations are in tree structure because one post is likely to spark multiple replying or reposting messages, our model allows the modeling of discourse roles in tree structure, which enables richer contexts to be captured. More details will be described in Section~\ref{ssec:msg-modeling}.

\subsection{Microblog Summarization}

Microblog summarization can be considered as a special case of text summarization, which is conventionally defined to discover essential content from given document(s), and produce concise and informative summaries covering important information~\cite{DBLP:journals/coling/RadevHM02}.
Summarization techniques can be generally categorized as extractive and abstractive methods~\cite{das2007survey}.
Extractive summarization captures and distills salient content, which are usually sentences, to form summaries.
Abstractive summarization focuses on identifying key text units, e.g., words and phrases, and then generates grammatical summaries based on these units.
Our summarization application falls into the category of extractive summarization.

Early work on microblog summarization attempts to apply conventional extractive summarization models directly, e.g., LexRank \cite{DBLP:journals/jair/ErkanR04}, The University of Michigan’s summarization system MEAD~\cite{DBLP:conf/lrec/RadevABBCDDHLLOQSTTWZ04}, TF-IDF \cite{DBLP:conf/socialcom/InouyeK11}, integer linear programming (ILP) \cite{liu2011sxsw,DBLP:conf/ecir/TakamuraYO11}, graph learning \cite{sharifi2010automatic}, etc.
Later, researchers have found that standard summarization models are not suitable on microblog posts because of the severe redundancy, noise, and sparsity problems exhibited in short and colloquial messages~\cite{DBLP:conf/wsdm/ChangWML13,DBLP:conf/emnlp/LiGWPW15}.
To solve these problems, one common solution is to use social signals such as the user influence and retweet counts to help summarization~\cite{DBLP:conf/coling/DuanCWZS12,DBLP:conf/coling/LiuLWZ12,DBLP:conf/wsdm/ChangWML13}.
Different from the aforementioned studies, we do not include external features such as the social network structure, which ensures the general applicability of our approach when applied to domains without such information.

Discourse has been reported useful to microblog summarization.
\citet{DBLP:journals/taslp/ZhangLGY13} and \citet{DBLP:conf/emnlp/LiGWPW15} leverage dialogue acts to indicate summary-worthy messages.
In the fields conversation summarization from other domains, e.g., meetings, forums, and emails, it is also popular to leverage the pre-detected discourse structure for summarization \cite{DBLP:conf/naacl/MurrayRCM06,DBLP:conf/acl/WangC13,DBLP:conf/emnlp/BhatiaBM14,DBLP:conf/cicling/McKeownSR07,DBLP:journals/nle/BokaeiSL16}.
\citet{DBLP:conf/sigdial/OyaC14} and \citet{DBLP:conf/acl/QinWK17} address discourse tagging together with salient content discovery on emails and meetings, and show the usefulness of their relations in summarization. For all the systems mentioned above, manually crafted tags and annotated data are required for discourse modeling. Instead, the discourse structure is discovered in a fully unsupervised manner in our model, which is represented by word distributions and can be deviated from any human designed discourse inventory. The effects of such discourse representations on salient content identification has never been explored in any previous work.

%% file: sections/model.tex
\section{The Joint Model of Conversational Discourse and Latent Topics}\label{sec:model}

We assume that the given corpus of microblog posts is organized as $C$ conversation trees based on reposting and replying relations.
Each tree $c$ contains $M_c$ microblog messages and each message $m$ has $N_{c,m}$ words in vocabulary. The vocabulary size is $V$. We separate three components, i.e., 
\textit{discourse}, \textit{topic}, and \textit{background}, underlying the given conversations, and use three types of word distributions to represent them.

At corpus level, there are $K$ topics represented by word distribution $\phi^T_k\sim Dir(\beta)$ ($k=1,2,...,K$). $\phi^D_d \sim Dir(\beta)$ ($d=1,2,...,D$) represents the $D$ discourse roles embedded in the corpus. In addition, we add a background word distribution $\phi^B \sim Dir(\beta)$ to capture general information (e.g., common words), which cannot indicate either discourse or topic.  $\phi^T_k$, $\phi^D_d$, and $\phi^B$ are all $V$-dimensional multinomial word distributions over the vocabulary. 
For each conversation tree $c$, $\theta_c\sim Dir(\alpha)$ models the mixture of topics and any message $m$ on $c$ is assumed to contain a single topic $z_{c,m}\in \{1,2,...,K\}$.
     
\subsection{Message-level Modeling}\label{ssec:msg-modeling}
     
For each message $m$ on conversation tree $c$, our model assigns two message-level multinomial variables to it, i.e., $d_{c,m}$ representing its discourse role and $z_{c,m}$ reflecting its topic assignment, whose definitions are given in turn in the following. 

\vspace{1em}

\noindent\textbf{Discourse Roles.} Our discourse detection is inspired by \citet{DBLP:conf/naacl/RitterCD10} that exploits the discourse dependencies derived from reposting and replying relations for assigning discourse roles. For example, a ``doubt'' message is likely to start controversy thus triggers another ``doubt'', e.g., [R5] and [R8] in Figure~\ref{fig:tree}.
Assuming that the index of $m$'s parent is $pa(m)$, we use transition probabilities $\pi_d \sim Dir(\gamma)$ ($d=1,2,...,D$) to explicitly model discourse dependency of $m$ to $pa(m)$. $\pi_d$ is a distribution over the $D$ discourse roles and $\pi_{d,d'}$ denotes the probability of $m$ assigned discourse $d'$ given the discourse of $pa(m)$ being $d$. Specifically, $d_{c,m}$ (discourse role of message $m$) is generated from discourse transition distribution $\pi_{d_{t,pa(m)}}$ where $d_{t,pa(m)}$ is the discourse assignment on $pa(m)$.
In particular, to create a unified generation story, we place a pseudo message emitting no word before the root of each conversation tree and assign dummy discourse indexing $D+1$ to it. $\pi_{D+1}$, the discourse transition from pseudo messages to roots, in fact models the probabilities of different discourse roles as conversation starter.

\vspace{1em}    

\noindent\textbf{Topic Assignments.} Messages on one conversation tree focus on related topics. To exploit such intuition in topic assignments, the topic of each message $m$ on conversation tree $c$, i.e., $z_{c,m}$, is sampled from the topic mixture $\theta_c$ of conversation tree $c$.
    
\subsection{Word-level Modeling}\label{ssec:word-modeling}
   
To distinguish varying types of word distributions to separately capture \textit{discourse}, \textit{topic}, and \textit{background} representations, we follow the solutions from previous work to assign each word as a discrete and exact source that reflects one particular type of word representation~\cite{DBLP:conf/acl/DaumeM06,DBLP:conf/naacl/HaghighiV09,DBLP:conf/naacl/RitterCD10}. To this end, for each word $n$ in message $m$ and tree $c$, a ternary variable $x_{c,m,n}\in \{\text{DISC},\text{TOPIC},\text{BACK}\}$ controls word $n$ to fall into one of the three types: \textit{discourse}, \textit{topic}, and \textit{background} word. 
In doing so, words in the given collection are explicitly separated into three types, based on which the word distributions representing discourse, topic, and background components are separated accordingly. 

\vspace{1em}
   
\noindent\textbf{Discourse words} (DISC) indicate the discourse role of a message, e.g., in Figure \ref{fig:tree}, ``How'' and the question mark ``?'' reflect that [R1] should be assigned the discourse role of ``question''. If $x_{c,m,n}=\text{DISC}$, i.e., $n$ is assigned as a discourse word, word $w_{c,m,n}$ is generated from discourse word distribution $\phi^D_{d_{c,m}}$, where $d_{c,m}$ is $m$'s discourse role. 

\vspace{1em}

\noindent\textbf{Topic words} (TOPIC) are the core topical words that describe topics being discussed in a conversation tree, such as ``Muslim'', ``order'', and ``Trump'' in Figure \ref{fig:tree}. When $x_{c,m,n}=\text{TOPIC}$, i.e., $n$ is assigned as a topic word, word $w_{c,m,n}$ is hence generated from the word distribution of the topic assigned to message $m$, i.e., $\phi^T_{z_{c,m}}$

\vspace{1em}

\noindent\textbf{Background words} (BACK) capture the general words irrelevant to either discourse or topic, such as ``those'' and ``of'' in Figure~\ref{fig:tree}. When word $n$ is assigned as a background word ($x_{c,m,n}=\text{BACK}$), word $w_{c,m,n}$ is then drawn from background distribution $\phi^B$.

\vspace{1em}
  
\noindent\textbf{Switching among Topic, Discourse, and Background.} We assume that messages of different
discourse roles may show different distributions of the word types as discourse, topic, and background. The ternary word type switcher $x_{c,m,n}$ is hence controlled by the the discourse role of message $m$. In specific, $x_{c,m,n}$ is drawn from the three-dimensional distribution $\tau_{d_{c,m,n}}\sim Dir(\delta)$ that captures the appearing probabilities of three types of words (DISC, TOPIC,
BACK), when the discourse assignment to $m$ is $d_{c,m,n}$, i.e., $x_{c,m,n}\sim Multi(\tau_{d_{c,m}})$. For instance, a statement message, e.g., [R3] in Figure~\ref{fig:tree}, may contain more content words for topic representation than a question to other users, e.g., [R1] in Figure~\ref{fig:tree}. In particular, stop words and punctuation are forced to be labeled as discourse or background words. By explicitly distinguishing different types of words with switcher $x_{c,m,n}$, we can thus separate the three types of word distributions that reflect discourse, topic, and background information.
    
\subsection{Generative Process and Parameter Estimation}

  \input{figures/topic-model}

In summary,  Figure \ref{fig:model} illustrates the graphical model of our generative process that jointly explores conversational discourse and latent topics. The following shows the detailed generative process of the conversation tree $c$:
    \begin{compactitem}
    	\item Draw topic mixture of conversation tree $\theta_c \sim Dir(\alpha)$
        \item For message $m=1$ to $M_c$
        \begin{compactitem}
        \item Draw discourse role $d_{c,m} \sim Multi(\pi_{d_{c,pa(m)}})$
        \item Draw topic assignment $z_{c,m} \sim Multi(\theta_c)$
		\item For word $n=1$ to $N_{c,m}$
		\begin{compactitem}
			\item Draw ternary word type switcher $x_{c,m,n}\sim Multi(\tau_{d_{c,m}})$
            \item If $x_{c,m,n}==\text{DISC}$
            \begin{compactitem}
				\item Draw $w_{t,s,n}\sim Multi(\phi^D_{d_{c,m}})$
			\end{compactitem}
            \item If $x_{c,m,n}==\text{TOPIC}$
			\begin{compactitem}
				\item Draw $w_{c,m,n}\sim Multi(\phi^T_{z_{c,m}})$
			\end{compactitem}
            \item If $x_{c,m,n}==\text{BACK}$
			\begin{compactitem}
				\item Draw $w_{c,m,n}\sim Multi(\phi^B)$
			\end{compactitem}
		\end{compactitem}
        \end{compactitem}
        \end{compactitem}
    
For parameter estimation, we use collapsed Gibbs Sampling \cite{griffiths2004finding} to carry out posterior inference for parameter learning. The hidden multinomial variables, i.e., message-level variables ($d$ and $z$) and word-level variable ($x$) are sampled in turn, conditioned on a complete assignment of all other hidden variables and hyper-parameters $\Theta=(\alpha, \beta, \gamma, \delta)$. For more details, we refer the readers to Appendix.

%% file: figures/topic-model.tex
 \begin{figure}
    	\centering
\includegraphics[width=9cm]{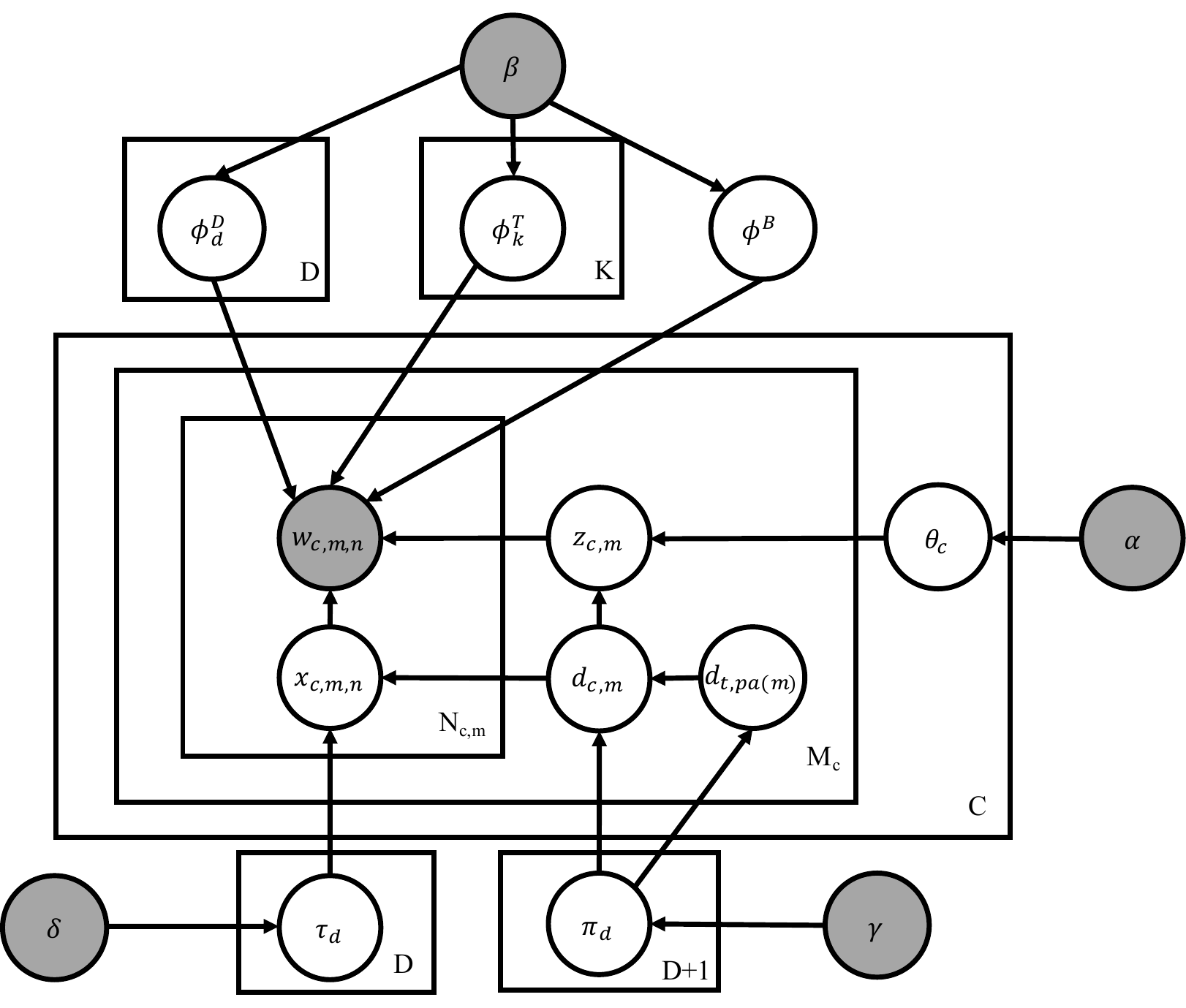}
		\caption{The graphical model illustrating the generative process of 
		our joint model of conversational discourse and latent topics.
		}\label{fig:model}
       \vskip-0.5em
	\end{figure}

%% file: sections/exp-set.tex
\section{Experiments on Topic Coherence} \label{sec:exp-coherence}

This section presents an experiment on the coherence of topics yielded by our joint model of conversational discourse and latent topics.

\subsection{Data Collection and Experiment Setup}\label{sect:data_and_setup}

\textbf{Datasets.} To examine the coherence of topics on diverse microblog datasets, we conduct experiments on datasets collected from two popular microblog websites: Twitter and Weibo\footnote{Weibo, short for Sina Weibo, is the biggest microblog platform in China and shares the similar market penetration as Twitter~\cite{rapoza2011china}. Similar to Twitter, it has length limitation of 140 Chinese characters}, where the messages are mostly in English and Chinese, respectively. Table \ref{tab:data} shows the statistics of our five datasets used to evaluate topic coherence. In the following, we give the details of their collection processes in turn.

\vspace{1em}

For Twitter data, we evaluate the coherence of topics on three datasets: \textit{SemEval}, \textit{PHEME}, and \textit{US Election}, and tune all models in our experiments on a large-scale development dataset from TREC2011 microblog track\footnote{\url{http://trec.nist.gov/data/tweets/}}.

\vspace{0.5em}
$\bullet$~\textit{SemEval}. We combine the data released for topic oriented sentiment analysis task in SemEval 2015\footnote{\url{http://alt.qcri.org/semeval2015/task10/}} and 2016\footnote{\url{http://alt.qcri.org/semeval2016/task4/}}.
To recover the missing ancestors in conversation trees, we use Tweet Search API to retrieve messages with the ``in-reply-to'' relations, and collect tweets in a recursive way until all the ancestors in a conversation are recovered.\footnote{Twitter search API: \url{https://developer.twitter.com/en/docs/tweets/search/api-reference/get-saved_searches-show-id}.
Twitter has allowed users to add comments in retweets (reposting messages on Twitter) since 2015, which enables that retweets to become part of a conversations.
In our dataset, the parents of $91.4$\% of such retweets can be recovered from the ``in reply to status id'' field returned by Twitter search API.}

\vspace{0.5em}

$\bullet$~\textit{PHEME}. This dataset is released by \citet{DBLP:journals/corr/ZubiagaLP16}, which contains conversations around rumors and non-rumors posted during five breaking events: \textit{Charlie Hebdo}, \textit{Ferguson}, \textit{Germanwings Crash}, \textit{Ottawa Shooting}, and \textit{Sydney Siege}.

\vspace{0.5em}

$\bullet$~\textit{US Election}. Considering that the \textit{SemEval} and \textit{PHEME} datasets cover relatively wide range of topics, we are interested in studying a more challenging problem: whether topic models can differentiate latent topics in a narrow scope.
To this end, we take political tweets as an example and conduct experiments on a dataset with Twitter discussions about the U.S. presidential election 2016. The dataset is extended from the one released by \citet{DBLP:conf/naacl/ZengLWBSW18} following three steps.
First, some raw tweets that are likely to be in a conversation are collected by searching conversation-type keywords via Twitter Streaming API\footnote{\url{https://developer.twitter.com/en/docs/tweets/filter-realtime/api-reference/post-statuses-filter.html}}, which samples and returns tweets matching the given keywords.\footnote{Conversation-type keywords are used to obtain tweets reflecting \textit{agreement}, \textit{disagreement}, and \textit{response}, which are likely to appear in Twitter conversations. Keyword list: \textit{agreement} -- ``agreed'', ``great point'', ``agree'', ``good point''; \textit{disagreement} -- ``wrong'', ``bad idea'', ``stupid idea'', ``disagree''; \textit{response} -- ``understand'', ``interesting'', ``i see''.}
Second, conversations are recovered via ``in-reply-to'' relations as what is done to build \textit{SemEval} dataset.
Third, the relevant conversations are selected where there exist at least one tweet containing election-related keywords.\footnote{The full list of election-related keywords: ``trump'', ``clinton'', ``hillary'', ``election'', ``president'', ``politics''.}

\vspace{1em}

\input{tables/data-statistics}

For Weibo data, We track the real-time trending hashtags\footnote{\url{ http://open.weibo.com/wiki/Trends/hourly?sudaref=www.google.com.hk&retcode=6102}} on Sina Weibo and use the hashtag-search API\footnote{\url{http://open.weibo.com/wiki/2/search/topics}} to crawl the posts matching the given hashtag queries. In the end, we build a large-scale corpus containing messages posted during Jan 2 to July 31, 2014. To examine the performance of models on varying topic distributions, we split the corpus into seven subsets, each containing messages posted in one month. We report the topic coherence on two randomly selected subsets, i.e., \textit{Weibo-1} and \textit{Weibo-2}. The remaining five datasets are used as development sets.

\vspace{1em}

\noindent\textbf{Comparisons.} 
Our model jointly identifies word clusters of discourse and topics, and explicitly explores their relations, i.e., the probabilities of different discourse roles in containing topical words (see Section \ref{ssec:word-modeling}), which is named as \textsc{topic+disc+rel} model in the rest of the article.
In comparison, we consider the following established models: 1) \textsc{LDA}: In this model, we consider each message as a document and directly apply latent Dirichlet model (LDA) \citep{DBLP:conf/nips/BleiGJT03,DBLP:journals/jmlr/BleiNJ03} on the collection. The implementation of \textsc{LDA} model is based on public toolkit GibbsLDA++.\footnote{\url{http://gibbslda.sourceforge.net/}}. 2) \textsc{BTM}: Biterm topic model (BTM)\footnote{\url{https://github.com/xiaohuiyan/BTM}} \cite{DBLP:conf/www/YanGLC13,DBLP:journals/tkde/ChengYLG14} is a state-of-the-art topic model for short texts. It directly models the topics of all word pairs (biterms) in each message, which has proven more effective on social media texts than LDA \cite{DBLP:conf/nips/BleiGJT03,DBLP:journals/jmlr/BleiNJ03}, one-topic-per-post Dirichlet multinomial mixture (DMM) \cite{DBLP:journals/ml/NigamMTM00}, and \citet{DBLP:conf/ecir/ZhaoJWHLYL11} (a DMM version on posts aggregated by authorship). According to the empirical study in \citet{DBLP:conf/acl/LiLGHW16}, \textsc{BTM} has a general better performance than a newer SATM model \cite{DBLP:conf/ijcai/QuanKGP15} on microblog data.

In particular, this article attempts to induce topics with little external resource. Therefore, we don't compare with either \citet{DBLP:conf/acl/LiLGHW16}, which depends on human annotation to train a discourse tagger, or topic models that exploit word embeddings \cite{DBLP:conf/acl/DasZD15,DBLP:journals/tacl/NguyenBDJ15,DBLP:conf/sigir/LiWZSM16,DBLP:journals/tois/LiDWZSM17,DBLP:conf/ijcai/XunLZGZ17,DBLP:conf/sigir/ShiLJSL17} pre-trained on large-scale external data. 
The external data in training embeddings should be in both the same domain and the same language of the given collection used for topic models, which limits the applicability of topic models in the scenarios without such data. Also, \citet{DBLP:conf/acl/LiLGHW16} has shown that topic models combining word embeddings trained on internal data give worse coherence scores than \textsc{BTM}, which has been considered in our comparison.

In addition to the existing models from previous work, we consider the following variants that explore topics by organizing messages as conversation trees: 

\vspace{0.5em}

$\bullet$~\textsc{topic only} model aggregates messages from one conversation tree as a pseudo-document, on which \citet{DBLP:conf/nips/ChemuduguntaSS06}, a model proven better than LDA in topic coherence, is used to induce topics on conversation aggregations, without modeling discourse structure. It involves a background word distribution to capture non-topic words, so as our \textsc{topic+disc+rel} model. However, different from our  \textsc{topic+disc+rel} model, the background word distribution is controlled by a general Beta prior without differentiating discourse roles of messages. 

\vspace{0.5em}

$\bullet$~\textsc{topic+disc} model is an extension to \citet{DBLP:conf/naacl/RitterCD10}, following which the switcher indicating a word as a discourse, topic, or background word are drawn from a conversation-level distribution over word types. Instead, in \textsc{topic+disc+rel}, word-type switcher depends on message-level discourse roles (shown in Section~\ref{ssec:word-modeling}).
In terms of topic generation of \textsc{topic+disc} model, as \citet{DBLP:conf/naacl/RitterCD10} is incapable of differentiating various latent topics, we follow the same procedure of \textsc{topic only} and \textsc{topic+disc+rel} model to draw topics from conversation-level topic mixture.
Another difference between \textsc{topic+disc} model and \citet{DBLP:conf/naacl/RitterCD10} is that the discourse roles of \textsc{topic+disc} are explored in tree-structured conversations while those in \citet{DBLP:conf/naacl/RitterCD10} are captured in context of conversation treads (paths of the conversation tree).

\vspace{1em}

\noindent\textbf{Hyper-parameters.} For the hyper-parameters of our joint \textsc{topic+disc+rel} model, we fix $\alpha=50/K$, $\beta=0.01$, following the common practice in previous work~\cite{DBLP:conf/www/YanGLC13,DBLP:journals/tkde/ChengYLG14}. For Twitter corpora, we set the count of discourse roles as $D=10$ according to previous setting in \citet{DBLP:conf/naacl/RitterCD10}. Since there is no analogue of $\gamma$ (controlling the prior for discourse role dependencies of children messages to their parents), $\delta$ (controlling the prior of distributions over topic, discourse, and background words given varying discourse roles), and discourse count $D$ in Chinese Weibo corpora, we tune them by grid search on development sets and obtain $\gamma=0.5$, $\delta=0.25$, and $D=6$ on Weibo data.

The hyper-parameters of \textsc{LDA} and \textsc{BTM} are set according to the best hyper-parameters reported in their original papers. For \textsc{topic only} and \textsc{topic+disc} model, the parameter settings are kept the same as \textsc{topic+disc +rel} model, since they are its variants. And the background switchers are parameterized by symmetric Beta prior on $0.5$, following the original setting from \citet{DBLP:conf/nips/ChemuduguntaSS06}. We run Gibbs samplings of all models with $1,000$ iterations to ensure convergence following \citet{DBLP:conf/ecir/ZhaoJWHLYL11,DBLP:conf/www/YanGLC13,DBLP:journals/tkde/ChengYLG14}

\vspace{1em}

\noindent\textbf{Preprocessing.} Before training topic models, we preprocess the datasets as follows. For Twitter corpora, we 1) filter non-English messages; 2) replace links, mentions (i.e., @username),
and hashtags with generic tags of ``URL'', ``MENTION'', and ``HASHTAG''; 3) tokenize messages and annotate part-of-speech (POS) tags to each word using Tweet NLP toolkit \cite{DBLP:conf/acl/GimpelSODMEHYFS11,DBLP:conf/naacl/OwoputiODGSS13}.\footnote{\url{http://www.cs.cmu.edu/~ark/TweetNLP/}}; 4) normalize all letters to lowercases.
For Weibo corpora, we 1) filter non-Chinese messages; 2) use FudanNLP toolkit~\cite{DBLP:conf/acl/QiuZH13} for word segmentation.
Then, for each dataset from Twitter or Sina Weibo, we generate a vocabulary and remove low-frequency words, i.e., words occurring less than five times.

For our \textsc{topic+disc+rel} model and its variants \textsc{topic only} and \textsc{topic+disc}  considering the conversation structure, we only remove digits but remain stop words and punctuation in the data because: 1) stop words and punctuation can be useful discourse indicators, such as the question mark ``?'' and ``what'' in indicating ``question'' messages; 2) these models are equipped with a background distribution $\phi^B$ to separate general information useless to indicate either discourse or topic, e.g., ``do'' and ``it''; 3) we forbid stop words and punctuation to be sampled as topical words by forcing their word type switcher $x\neq $TOPIC in word generation (shown in Section~\ref{ssec:word-modeling}). For \textsc{LDA} and \textsc{BTM} that cannot separate non-topic information, we filter out stop words and short messages with less than two words in preprocessing, which keeps the same as their common settings to ensure comparable performance.\footnote{We also conducted evaluations on the \textsc{LDA} and \textsc{BTM} versions without this pre-processing step, and they gave worse coherence scores.}

\vspace{1em}

\noindent\textbf{Evaluation Metrics.} Topic model evaluation is inherently difficult. Though in many previous studies, perplexity is a popular metric to evaluate the predictive abilities of topic models given held-out dataset with unseen words~\cite{DBLP:journals/jmlr/BleiNJ03}, we don't consider perplexity here because high perplexity does not necessarily indicate semantically coherent topics in human perception~\cite{DBLP:conf/nips/ChangBGWB09}.

The quality of topics is commonly measured by UCI~\cite{DBLP:conf/naacl/NewmanLGB10} and UMass coherence scores~\cite{DBLP:conf/emnlp/MimnoWTLM11}, assuming that words representing a coherent topic are likely to co-occur within the same document. We only consider UMass coherence here as UMass and UCI generally agree with each other according to~\citet{DBLP:conf/emnlp/StevensKAB12}.
We also consider a newer evaluation metric CV coherence measure~\cite{DBLP:conf/wsdm/RoderBH15}, as it has been proven to give the scores closest to human evaluation compared with other widely used topic coherence metrics,
including UCI and UMass scores.\footnote{\url{http://aksw.org/Projects/Palmetto.html}} For CV coherence measure, in brief, given a word list for topic representations, i.e., the top $N$ words by topic-word distribution, some known topic coherence measures are combined, which estimates of how similar their co-occurrence patterns with other words are in context of a sliding window from Wikipedia. 

%% file: tables/data-statistics.tex
\begin{table}
\begin{center}
\caption{Statistics of our five datasets on Twitter and Sina Weibo for evaluating topic coherence.}\label{tab:data}
\begin{tabular}{|lrrr|}
\hline
DataSet	&	\# of trees	&	\# of messages	&	Vocab size\\
\hline
\hline
\textbf{\underline{Twitter}} & & &\\
\textit{SemEval} & 8,652 & 13,582 & 3,882\\
\textit{PHEME}	& 7,961 & 92,883 & 10,288\\
\textit{US Election} & 4,396 & 33,960 & 5,113\\
\hline
\hline
\textbf{\underline{Weibo}} & & &\\
\textit{Weibo-1}	&	9,959	&	9,1268	&	11,849\\
\textit{Weibo-2}	&	21,923	&	277,931	&	19,843\\
\hline	
\end{tabular}
\end{center}
\end{table}

%% file: sections/exp-res.tex
\subsection{Main Comparison Results}

\input{tables/umass_results.tex}

We evaluate topic models with two sets of $K$, i.e., the number of topics, $K=50$ and $K=100$, following previous settings \cite{DBLP:conf/acl/LiLGHW16}. Table \ref{tab:umass} and Table \ref{tab:cv} show the UMass and CV scores for topics produced on the evaluation corpora, respectively. For UMass coherence, the top $5$, $10$, $15$, and $20$ words of each topic are selected for evaluation. For CV coherence, the top $5$ and $10$ words are selected. \footnote{Palmetto toolkit only allows at most $10$ words as input for CV score calculation.} Note that we cannot report CV scores on Chinese Weibo corpora since CV coherence is calculated based on a Wikipedia dataset, which does not have Chinese version so far. 
From the results, we have the following observations:

\vspace{0.5em}

\noindent $\bullet$~\textbf{\textit{Conventional topic models cannot perform well on microblog messages.}} From all the comparison results, the topic coherence given by \textsc{LDA} is the worst, which may because of the sparseness of document-level word concurrence patterns in short posts.

\vspace{0.5em}

\noindent $\bullet$~\textbf{\textit{Considering conversation structure is useful to topic inference.}} Using the contextual information provided by conversations, \textsc{topic only} model produced competitive
results compared to the state-of-the-art \textsc{BTM}
model on short text topic modeling. This observation indicates the effectiveness of using the conversation structure to enrich context and thus results in latent topics of reasonably good quality.

\vspace{0.5em}

\noindent $\bullet$~\textbf{\textit{Jointly learning discourse information helps produce coherent topics.}} \textsc{topic+disc} and \textsc{topic+disc+rel} models yield generally better coherence scores than \textsc{topic only} model, which explores topics without considering discourse. The reason may be that, additionally exploring discourse in non-topic information helps recognize non-topic words, which further facilitates the separation of topical words from non-topic ones.

\vspace{0.5em}

\noindent $\bullet$~\textbf{\textit{Considering discourse roles of messages in topical word generation is useful.}} 
The results of \textsc{topic+disc+rel} are the best in most settings. One important reason is that \textsc{topic+disc+rel} model explicitly explores the different probabilities of messages with varying discourse roles in containing topical or non-topic words, while the other models separate topical content from non-topic information regardless of the different discourse roles of messages. 
This observation demonstrates that messages with different discourse roles do vary in tendencies to cover topical words, which provides useful clues for key content words to be identified for topic representation.

\input{tables/cv_results.tex}

%% file: tables/umass_results.tex
\begin{table}\scriptsize

\begin{center}
\caption{UMass coherence scores for topics produced by various models. Higher is better. K50: 50 topics; K100: 100 topics; N: the number of top words ranked by topic-word probabilities. W/o conversation: the models consider each message as a document and infers topics without using conversational information. W/ conversation: messages first are organized as conversation trees and then the model induce topics in context of conversation trees. Higher scores indicate better coherence. Best result for each setting is in \textbf{bold}.}
\label{tab:umass}
\scalebox{0.88}{
\begin{tabular}{|l|l|rr|rr|rr|rr|rr|}
\hline
\multirow{2}{*}{$N$}	&	
\multirow{2}{*}{Model}	&	
\multicolumn{2}{c|}{\textit{Weibo-1}}	&	\multicolumn{2}{c}{\textit{Weibo-2}}	& \multicolumn{2}{|c}{\textit{SemEval}}&\multicolumn{2}{|c|}{\textit{PHEME}}&\multicolumn{2}{|c|}{\textit{US Election}}\\
   \cline{3-12}
&&	K50	&	K100	&	K50	&	K100	&	K50	&	K100	&	K50	&	K100&K50	&	K100 	\\
\hline
\hline
\multirow{7}{*}{5}		
&\textbf{\underline{W/o conversation}}	& & & & & & & & &&\\
&\textsc{LDA}	&	-11.77	&	-11.57	&	-10.56	&	-12.08&-12.20&-12.02 &-13.27&-13.98& \textbf{-10.07}&-10.89\\
&\textsc{BTM}	&	-9.56	&	-8.74	&	-8.65 & -9.88 & -9.93 & -9.61 &-10.22&-10.44& -12.15 & -12.15\\
&\textbf{\underline{W/ conversation}}	& & & & & & & &&& \\
&\textsc{topic only}	&	\textbf{-8.00}	&	-8.78 &	-9.45	&	-10.06 & -8.93 & -8.88 &-10.82&-10.63& -10.75 & -10.98\\
&\textsc{topic+disc}	&	-9.47	&	-8.87	& -9.85	&	\textbf{-9.60}&\textbf{-8.42}& -8.26 &-10.54&\textbf{-10.40}& -10.36 & -11.17\\
&\textsc{topic+disc+rel}	&	-8.53	&	\textbf{-8.66}	&	\textbf{-8.00}	&	-9.84 & -8.47 & \textbf{-8.19} &\textbf{-10.21}&-10.41& -11.14& \textbf{-10.75}\\
\hline
\hline
\multirow{7}{*}{10}	
&\textbf{\underline{W/o conversation}}	&&& & & & & & & & \\
&\textsc{LDA}	&	-120.06	&	-123.74	&	-117.00	&	-123.98&-128.15&-132.96&-138.99&-145.44&-105.21&-110.82\\
&\textsc{BTM}	&	-89.98	&	-86.96	& -87.97 & \textbf{-93.03} & -105.76 & -105.98 &-108.70&-111.32 & -114.85 &-123.42\\
&\textbf{\underline{W/ conversation}}	& &&& & & & & & & \\	
& \textsc{topic only}	&	-90.53	&	-89.89	& 	-108.51	&	-101.20&-89.02&-90.53&-105.62&-108.25&\textbf{-104.29}&-108.51\\
&\textsc{topic+disc}	&	-91.96	&	-88.75	&		-100.77	&	-100.58 & -87.89 & -91.82&-106.58&\textbf{-107.14}&-105.21&-108.31\\
&\textsc{topic+disc+rel}	&	\textbf{-86.91}	&	\textbf{-87.05}	&	\textbf{-83.59}	&	-98.19 & \textbf{-86.48} & \textbf{-90.02} &\textbf{-105.27}&-107.91& -104.99&\textbf{-107.03}\\
\hline
\hline
\multirow{7}{*}{15}	
&\textbf{\underline{W/o conversation}}	& & & & & &&& & & \\
&\textsc{LDA}	&	-367.68	&	-357.88	&	-366.31	&	-373.98&-383.05&-391.67&-418.58&-424.66&-429.89&-436.87\\
&\textsc{BTM}	&	-265.80	&	-262.62	&	-281.06 & \textbf{-281.46} & -307.23 & -323.37&-328.36&-339.94&-344.99&-360.95\\
&\textbf{\underline{W/ conversation}}	& & & & & & & & &&\\	
&	\textsc{topic only}	&	-261.98 & -260.62	&	-298.77	&	-294.51&-257.92&-266.86&--313.96&\textbf{-315.78}&-313.07&-319.99\\
&\textsc{topic+disc}	&	-261.30	&	-259.23	&	-301.99	& -293.21 & -261.25 & -265.88&-313.22&-320.05&-317.14&-317.82\\
&\textsc{topic+disc+rel}	&	\textbf{-254.94}	&	\textbf{-256.47}	&	\textbf{-249.32}	&	-287.82 & \textbf{-256.83} & \textbf{-265.71}&\textbf{-312.49}&-319.01&\textbf{-312.90}&\textbf{-315.59}\\
\hline
\hline
\multirow{7}{*}{20}
&\textbf{\underline{W/o conversation}}	& & & & & & & & &&\\
&\textsc{LDA}	&	-771.34	&	-736.55	&	-718.48	&	-741.77	&	-777.00&-782.51&-856.37&-859.59&-898.77&-892.84\\
&\textsc{BTM}	&	-559.69	&	-553.62	&	-526.01	&	-586.65 & -636.15 & -669.16&-682.39&-709.81&-713.05&-739.35\\
&\textbf{\underline{W/ conversation}}	& & & & & & & & &&\\	
&	\textsc{topic only}	&	-528.13	&	-527.71	&	-602.16	&	-597.80&\textbf{-529.39}&-541.31&-643.91&\textbf{-647.74}&-634.97&-638.10\\
&\textsc{topic+disc}	&	-530.23	&	-524.15	&	-607.84	&	-585.99 & -535.22&-541.18&-641.82&-656.16&-641.77&-639.35\\
&\textsc{topic+disc+rel}	&	\textbf{-518.97}	&	\textbf{-519.11} &	\textbf{-509.79} & \textbf{-578.80} & -530.56 & \textbf{-538.31}&\textbf{-637.18}&-650.70&\textbf{-629.42}&\textbf{-634.22}\\
\hline
\end{tabular}
}
\end{center}
\end{table}

%% file: tables/cv_results.tex
\begin{table}
\caption{CV coherence scores for topics produced by various models on Twitter. Higher is better. K50: 50 topics; K100: 100 topics; N: the number of top words ranked by topic-word probabilities. Higher scores indicate better coherence. Best result for each setting is in \textbf{bold}.}\label{tab:cv}
\label{tab:coherence}
\begin{center}
\begin{tabular}{|l|l|ll|ll|ll|}
\hline
\multirow{2}{*}{$N$}	&	
\multirow{2}{*}{Model}	&	
\multicolumn{2}{c|}{\textit{SemEval}} &\multicolumn{2}{c|}{\textit{PHEME}}& \multicolumn{2}{c|}{\textit{US Election}}\\
   \cline{3-8}
&&	K50	&	K100 & K50 & K100&K50&K100\\
\hline
\hline
\multirow{ 7}{*}{5}		
&\textbf{\underline{W/o conversation}}&&&&&&\\
& \textsc{LDA} & 0.514 & 0.498 &0.474&0.470& 0.473 & 0.470\\
& \textsc{BTM} & 0.528 & 0.518 &0.486&0.477& 0.481 & 0.480\\
&\textbf{\underline{W/ conversation}} &&&&&&\\
&\textsc{topic only}	&	0.526	&	0.521 & \textbf{0.492} &0.485&0.477& 0.475\\
&\textsc{topic+disc}	&	0.526 & 0.523 & 0.481 &0.483&0.475& 0.478\\
&\textsc{topic+disc+rel}	&	\textbf{0.535} & \textbf{0.524} &0.491&\textbf{0.493}& \textbf{0.482} & \textbf{0.483}\\
\hline
\hline
\multirow{7}{*}{10}
&\textbf{\underline{W/o conversation}}&&&&&&\\
& \textsc{LDA} & 0.404 & 0.401 &0.375&0.378& 0.351&0.359\\
& \textsc{BTM} & 0.412 & 0.406 &0.386&0.385& 0.354&0.363\\
&\textbf{\underline{W/ conversation}} & & &&&&\\		
&\textsc{topic only}	&	0.399	&	\textbf{0.410} &0.388& 0.385& 0.359 &0.360\\
&\textsc{topic+disc}	&	0.408 & \textbf{0.410} &0.388&\textbf{0.386}& 0.356 &0.364\\
&\textsc{topic+disc+rel}	&	\textbf{0.414} & \textbf{0.410}&\textbf{0.398}&\textbf{0.386}&\textbf{0.366}&\textbf{0.366}\\
\hline
\end{tabular}
\end{center}
\end{table}

%% file: sections/case-study.tex
\input{tables/topic-case.tex}

\subsection{Case Study}
To further evaluate the interpretability of the latent topics and discourse roles learned by our \textsc{topic+disc+rel} model, we present a qualitative analysis on the output samples. 

\vspace{1em}

\noindent\textbf{Sample Latent Topics.}
We first present a qualitatively study on the sample produced topics. Table \ref{tab:case-study} displays the top 15 words of topic ``Trump is a raciest'' induced by different models on US election dataset given $K=100$.\footnote{If there are multiple latent topics related to ``Trump is a raciest'', we pick up the most relevant one and display its representative words.} We have the following observations:

\vspace{0.5em}

\noindent$\bullet$~It is challenging to extract coherent and meaningful topics from short and informal microblog messages. Without employing an effective strategy to alleviate the data sparsity problem, \textsc{LDA} mixes the generated topic with \textit{non-topic words}\footnote{\textit{Non-topic words} cannot clearly indicate the corresponding topic. Such words can occur in  messages covering very different topics. For example, in Table \ref{tab:case-study}, the word ``opinion'' is a non-topic word for ``Trump is a racist'', because an ``opinion'' can be voiced on diverse people, events, entities, etc. }, such as ``direct'', ``describe'', ``opinion'', etc., which are also likely to appear in messages whose topics are very different from ``Trump is a raciest''.

\vspace{0.5em}

\noindent$\bullet$~By aggregating messages based on conversations, \textsc{topic only} model yields the topic competitive to the one produced by state-of-the-art \textsc{BTM} model. The reason behind this observation could be that the conversation context provides rich word co-occurrences patterns in topic induction, which is beneficial to alleviate the data sparsity.

\vspace{0.5em}

\noindent$\bullet$~The topics produced by \textsc{topic+disc} and \textsc{topic+disc+rel} model contain less non-topic words than \textsc{topic only} model, which does not consider discourse information when generating topics, and thus contains many general words, such as ``thing'' and ``work'', which cannot clearly indicate ``Trump is a raciest''.

\vspace{0.5em}

\input{tables/discourse-case.tex}

\noindent$\bullet$~The topic generated by \textsc{topic+disc+rel} best describes the topic ``Trump is a racist'' except for a non-topic word ``call'' at the end of the list. This is because it successfully discovers messages with discourse roles that are more likely to cover words describing the key focus in the conversations centering on ``Trump is a racist''. Without capturing such information, the topic produced by \textsc{topic+disc} model contains some non-topic words like ``yeah'' and ``agree''.

\vspace{1em}

\noindent\textbf{Sample Discourse Roles.} To show the discourse representation exploited by our \textsc{topic+disc+rel} model, we then present the sample discourse roles learned from PHEME dataset in Table~\ref{tab:discourse-case}. Although this is merely a qualitative human judgment, there appears to be interesting word clusters that reflect varying discourse roles found by our model without the guidance from any manual annotation on discourse.
In the first column of Table~\ref{tab:discourse-case}, we intuitively name the sample generated discourse roles, which are based on our interpretations of the word cluster, and are provided to benefit the reader. In below we discuss each displayed discourse role in turn:

\vspace{0.5em}

\noindent$\bullet$~\textit{Statement} presents arguments and judgments, where words like ``should'', ``need'' are widely used in suggestions and ``if'' occurs when conditions are given.

\vspace{0.5em}

\noindent$\bullet$~\textit{Reaction} expresses non-argumentative opinions. Compared to ``statement'' messages, ``reaction'' messages are straightforward and generally does not contain detailed explanations (e.g., conditions). Examples include simple feeling expressions, indicated by ``oh'', ``!!!'' and acknowledgements, indicated by ``thank'', ``thanks''. 

\vspace{0.5em}

\noindent$\bullet$~\textit{Question} represents users asking questions to other users, implied by the question mark ``?'', ``what'', ``why'', etc. 

\vspace{0.5em}

\noindent$\bullet$~\textit{Doubt} expresses strong opinions against something. Example indicative words are ``but'', ``don't'', ``just'', the question mark ``?'', etc.

\vspace{0.5em}

\noindent$\bullet$~\textit{Reference} is for quoting external resource, which is implied by words like ``from'', colon, and quotation marks. The usage of hashtags\footnote{On Twitter, a hashtag serves as a special URL, which can link other messages sharing the same hashtag.} and URLs are also prominent.

%% file: tables/topic-case.tex
\begin{table}
\begin{center}
\caption{The extracted topics describing ``Trump is a raciest''. Top $15$ words are selected by likelihood. Words are listed in decreasing order of probability given the topic. The words in red and boldface indicate non-topic words.}\label{tab:case-study}
\begin{tabular}{|m{12.5cm}|}
\hline
\underline{\textbf{W/o conversation}}\\
\textsc{LDA}: \\
\textcolor{red}{\textbf{call}} racist democracy \textcolor{red}{\textbf{opinion}} race racism ignorant \textcolor{red}{\textbf{definition}} bigot \textcolor{red}{\textbf{direct}} mexican entitle \textcolor{red}{\textbf{describe}} card bigotry\\
\textsc{BTM}: \\
racist people hate trump white racism muslims \textcolor{red}{\textbf{agree}} race \textcolor{red}{\textbf{call}} group \textcolor{red}{\textbf{make}} fact problem immigrant\\
\hline
\hline
\underline{\textbf{W/ conversation}}\\
\textsc{topic only}: \\
people black white \textcolor{red}{\textbf{understand}} racist \textcolor{red}{\textbf{vote}} agree wrong \textcolor{red}{\textbf{work}} president \textcolor{red}{\textbf{thing}} trump privilege disagree \textcolor{red}{\textbf{system}}\\
\textsc{topic+disc}:\\
white racist \textcolor{red}{\textbf{yeah}} trump privilege black race \textcolor{red}{\textbf{agree}} people bias \textcolor{red}{\textbf{true}} state issue
\textcolor{red}{\textbf{understand}} muslims\\
\textsc{topic+disc+rel}:\\
white people black racist hate race wrong privilege america muslims trump kill racism illegal \textcolor{red}{\textbf{call}}\\
\hline
\end{tabular}
\end{center}
\end{table}

%% file: tables/discourse-case.tex
\begin{table}
\begin{center}
\caption{Top $30$ representative terms for sample discourse
roles discovered by our \textsc{topic+disc+rel} model in PHEME dataset given $K=100$. Names of
the discourse roles are our interpretations according to the generated word distributions.}\label{tab:discourse-case}
\begin{tabular}{|c|m{10.5cm}|}
\hline
\textit{Statement}&
MENTION . the they are HASHTAG we , to and of in them all their ! be will these our who \& should do this for if us need have\\
\hline
\textit{Reaction}&MENTION ! . you URL HASHTAG for your thank this , on my i thanks so ... and a !! the are me please oh all very !!! - is\\
\hline
\textit{Question}&MENTION ? you the what are is do HASHTAG why they that how this a to did about who in he so or u was it know can does on\\
\hline
\textit{Doubt}&MENTION . you i , your a are to don't it but that if know u not me i'm and do have my think ? you're just about was it's\\
\hline
\textit{Reference}&: URL MENTION HASHTAG in `` '' . at , the of on a - has from is to after and are " been have as more for least 2\\
\hline
\end{tabular}
\end{center}
\end{table}

%% file: sections/summ.tex
\section{Downstream Application: Conversation Summarization on Microblogs}\label{exp-summ}

Section \ref{sec:exp-coherence} has shown that conversational discourse is helpful to recognize key topical information from short and informal microblog messages. We are hence interested in whether the induced topic and discourse representations can also benefit downstream applications. Here we take microblog summarization as an example, which suffers from the data sparsity problem \cite{DBLP:conf/wsdm/ChangWML13,DBLP:conf/emnlp/LiGWPW15}, similar to topic modeling on short texts. In this article, we focus on a subtask of microblog summarization, i.e., \textbf{microblog conversation summarization}, and present an empirical study to show how our output can be used to predict critical content in conversations.

We first present the task description. Given a conversation tree, succinct summaries should be produced by extracting salient content from the massive reposting and replying messages in the conversation. It helps users understand the key focus of a microblog conversation. It is also named as microblog context summarization in some previous work~\citep{DBLP:conf/wsdm/ChangWML13,DBLP:conf/emnlp/LiGWPW15}, because the produced summaries captures informative content in the lengthy conversations and provide valuable contexts to a short post, such as the background information and public opinions. 
In this task, the \textit{input} is a microblog conversation tree, such as the one shown in Figure \ref{fig:tree}, and the \textit{output} is a subset of replying or reposting messages covering salient content of the input post. 

\subsection{Data Collection and Experiment Setup}\label{ssec:summ-set}

We then conduct an empirical study on the outputs of our joint model on microblog conversation summarization, whose the data preparation and setup processes are presented in this section.

\vspace{1em}
\noindent\textbf{Datasets.} Our experiments are conducted on a large-scale corpus containing ten big conversation trees collected from Sina Weibo, which is released by our prior work \citet{DBLP:conf/emnlp/LiGWPW15} and constructed following the settings described in \citet{DBLP:conf/wsdm/ChangWML13}.
The conversation trees discuss hot events taking place during January 2nd -- July 28th 2014, and are crawled using PKUVIS toolkit~\citep{DBLP:conf/apvis/RenZWLY14}. Each conversation tree has more than $12$K messages on average and covers discussions about social issues, breaking news, jokes, celebrity scandals, love, and fashion, which matches the official list of typical categories for microblog posts released by Sina Weibo.\footnote{\url{d.weibo.com/}} For each conversation tree, three experienced editors are invited to write summaries. Based on the manual summaries written by them, we conduct ROUGE evaluation, shown in Section \ref{ssec:rouge-eval}.

\input{tables/summ_statistics.tex}

Though compared with many other tasks in NLP and IR community, the corpus looks relatively small. However, to the best of our knowledge, it is the only publicly available dataset for conversation summarization so far.\footnote{The corpus of \citet{DBLP:conf/wsdm/ChangWML13} is not publicly available.} Because it is essentially difficult and time-consuming for human editors to write summaries for conversation trees ascribed to their massive nodes and complex structure \cite{DBLP:conf/wsdm/ChangWML13}. The editors could hardly reconstruct the conversation trees though they go through all the message nodes. In the evaluation for each tree, we compute the average ROUGE F1 score between the model-generated summary and the three human-generated summaries.

\vspace{1em}

\noindent\textbf{Summary Extraction.} Here we describe how summaries are produced given the outputs of topics models. For each conversation tree $c$, given the latent topics produced by topic models, we use a content word distribution $\gamma_c$ to describe its core focus and topic. Eq. \ref{eq:content-distribution} shows the formula to compute $\gamma_c$.

\begin{equation}\label{eq:content-distribution}
\gamma_{c,v} = Pr(v\text{ is a TOPIC word in }c)=\sum_{k=1}^K\theta_{c,k}\cdot\phi^T_{k,v}
\end{equation}
We further plug in $\gamma_c$ to the criterion proposed by~\citet{DBLP:conf/naacl/HaghighiV09}. The goal is to extract $L$ messages to form a summary set $E_c^*$ that closely matches $\gamma_c$. In our joint model, salient content of tree $c$ is captured without including background noise (modeled with $\phi^B$) or discourse indicative words (modeled with $\delta_d^D$).
Following~\citet{DBLP:conf/naacl/HaghighiV09}, conversation summarization is cast into the following Integer Programming (IP) problem:

    \begin{equation}\label{eq:msg-ext}
		E_c^*=\arg\min_{|E_c|= L} KL(\gamma_c||U(E_c))
	\end{equation}
where $U(E_c)$ denotes the empirical unigram distribution of the candidate summary set $E_c$ and $KL(P||Q)$ is the Kullback-Lieber (KL) divergence defined as $\sum_w P(w)\log \frac{P(w)}{Q(w)}$.\footnote{To ensure the value of KL-divergence to be finite, we smooth $U(E_c)$ with $\beta$, which also serves as the smoothing parameter of $\phi^T_k$ (Section \ref{sec:model}).} 
In implementation, as globally optimizing Eq. \ref{eq:msg-ext} is exponential in the total number of messages in a conversation, which is a non-deterministic polynomial-time (NP) problem, we use the greedy approximation adopted in \citet{DBLP:conf/naacl/HaghighiV09} for local optimization. Specifically, messages are greedily added to a summary so long as they minimize the KL-divergence in the current step.

\vspace{1em}

\noindent\textbf{Comparisons.} We consider baselines that rank and select messages by (1) \textsc{length}; (2) \textsc{popularity} (\# of reposts and replies); (3) \textsc{user} influence (\# of authors' followers); (4) message-message text similarities using \textsc{LexRank}~\cite{DBLP:journals/jair/ErkanR04}.
We also consider two state-of-the-art summarizers in comparison: 1) \textsc{\citet{DBLP:conf/wsdm/ChangWML13}}, a fully \textit{supervised} summarizers with manually crafted features; 2) \textsc{\citet{DBLP:conf/emnlp/LiGWPW15}}, a random walk variant summarizer incorporating outputs of \textit{supervised} discourse tagger.
In addition, we compare the summaries extracted based on the topics yielded by our \textsc{topic+disc+rel} model with those based on the outputs of its variants, i.e., \textsc{topic only} and \textsc{topic+disc} model. 

\vspace{1em}

\noindent\textbf{Preprocessing.} For baselines and the two state-of-the-art summarizers, we filter out non-Chinese characters in preprocessing step following their common settings.\footnote{We have also conducted evaluations on the versions without this pre-processing step, and they gave worse ROUGE scores.}
For summarization systems based on our topic model variants, i.e., \textsc{topic only}, \textsc{topic+disc}, and \textsc{topic+disc+rel}, the hyper-parameters and preprocessing steps keeps the same as Section \ref{sect:data_and_setup}.

\subsection{ROUGE Comparison}\label{ssec:rouge-eval}

\input{tables/rouge_results.tex}
\input{tables/human_results.tex}
\input{tables/summ_case.tex}

We quantitatively evaluate the performance of summarizers using ROUGE scores ~\cite{lin2004rouge} as benchmark, a widely used standard for automatic summarization evaluation based on the overlapping units between a produced summary and a gold-standard reference. In specific, Table \ref{tab:summ} reports ROUGE-1, ROUGE-2, ROUGE-L, and ROUGE-SU4 output by ROUGE 1.5.5.\footnote{\url{github.com/summanlp/evaluation/tree/master/ROUGE-RELEASE-1.5.5}. Note that the absolute scores of comparison models here are different from those reported in \citet{DBLP:conf/emnlp/LiGWPW15}. Because the ROUGE scores reported here are given
by ROUGE 1.5.5, while \citet{DBLP:conf/emnlp/LiGWPW15} uses Dragon toolkit \cite{DBLP:conf/ictai/ZhouZH07} for ROUGE calculation. Despite of the difference in absolute scores, the
trends reported here remain the same with \citet{DBLP:conf/emnlp/LiGWPW15}}. From the results, we can observe that:

\vspace{0.5em}
\noindent$\bullet$~\textbf{\textit{Simple features are not effective for summarization.}} The poor performance of all baselines demonstrates that microblog summarization is a challenging task. It is not possible to trivially rely on simple features such as length, message popularity, user influence, or text similarities to identify summary-worthy messages because of the colloquialness, noise, and redundancy exhibited in microblog texts.

\vspace{0.5em}

\noindent$\bullet$~\textbf{\textit{Discourse can indicate summary-worthy content.}} The summarization system based on \textsc{topic+disc+rel} model has generally better ROUGE scores than \textsc{topic+disc} based system. 
It also yields competitive and even slightly better results than \citet{DBLP:conf/emnlp/LiGWPW15}, which relies on a supervised discourse tagger. These observations demonstrate that \textsc{topic+disc+rel} model, without requiring gold-standard discourse annotation, is able to discover the discourse roles that are likely to convey topical words, which further reflect salient content for conversation summarization.

\vspace{0.5em}

\noindent$\bullet$~\textbf{\textit{Directly applying the outputs of our joint model of discourse and topics to summarization might not be perfect.}} In general, \textsc{topic+disc+rel} based system achieves the best F1 scores in ROUGE comparison, which implies that the yielded discourse and topic representations can somehow indicate summary-worthy content, although large margin improvements
are not observed. In Section~\ref{ssec:error-analysis}, we will analyze the errors and present a potential solution for further improving summarization results.

\subsection{Human Evaluation Results}\label{ssec:human-eval}

To further evaluate the generated summaries, we conduct human evaluations on informativeness (Info), conciseness (Conc) and readability (Read) of the extracted summaries. Two native Chinese speakers are invited to read the output summaries and subjectively rate on a 1-5 Likert scale and in 0.5 units, where a higher rating indicates better quality. Their overall inter-rater agreement achieves Krippendorff's $\alpha$ of $0.73$, which indicates reliable results~\cite{krippendorff2004content}. Table \ref{tab:rate} shows the average ratings by the two raters and over ten conversation trees.

As can be seen, despite of the closing results produced by supervised and well-performed unsupervised systems in automatic ROUGE evaluation (shown in Section~\ref{ssec:rouge-eval}), when the outputs are judged by humans, supervised systems \textsc{\citet{DBLP:conf/wsdm/ChangWML13}} and \textsc{\citet{DBLP:conf/emnlp/LiGWPW15}}, with supervision on summarization and discourse respectively, achieve much higher ratings than unsupervised systems on all the three criterion. This observation demonstrates that microblog conversation summarization is essentially challenging, where manual annotations, although with high cost in time and efforts involved, can provide useful clues in guiding systems to produce summaries that will be liked by humans. 
Particularly, the ratings given to \textsc{\citet{DBLP:conf/emnlp/LiGWPW15}} are higher than all systems in comparison by large margins, which indicates that the human-annotated discourse can well indicate summary-worthy content and also confirms the usefulness of considering discourse in microblog conversation summarization. 

Among unsupervised methods, the summarization results based on our \textsc{topic+disc+rel} model achieves generally better ratings than other comparison methods. The possible reasons are: 1) When separating topic words from discourse and background words, it also filters out irrelevant noise and distills important content; 2) It can exploit the tendencies of messages with varying discourse roles in containing core content, thus is able to identify ``bad'' discourse roles that bring redundancy or irrelevant words, which disturbs reading experience.

To further analyze the generated summaries, we conduct a case study and display a sample summary summary generated based on \textsc{topic+disc+rel} model in
Table \ref{tab:summary}. In this case, the input conversation is about the sexism issue in Chinese college entrance. As we can see, the produced summary covers salient comments that is helpful in understanding public opinions towards the gender discriminative problem. However, taking a closer look at the produced summaries, we observe that the system selects messages that contain sentiment-only information, such as ``Good point! I have to repost this!'', and therefore affect the quality of the generated summary. The observation from this summary case suggest that, in addition to discourse and background, sentiment component should be effectively captured and well separated for further improving the summarization results. The potential extension of the current summarization system to additionally incorporate sentiment will be discussed in Section~\ref{ssec:error-analysis}.

\subsection{Error Analysis and Further Discussions} \label{ssec:error-analysis}

Taking a closer look at our produced summaries, one major
source of incorrect selection of summary-worthy messages is based on the fact that sentiment is prevalent on microblog conversations, such as ``love'' in [R5] and ``poor'' in [R6] of Figure \ref{fig:tree}.
Without an additional separation of sentiment-specific information, the yielded topic representations might be mixed with sentiment components. For example, in Table \ref{tab:case-study}, the topic generated by \textsc{topic+disc+rel} model contains sentiment words like ``wrong'' and ``hate''. Therefore, a direct use of the topic representations to extract summaries will unavoidably select messages that mostly reflect sentimental component, which is also illustrated by the case study in Section~\ref{ssec:human-eval}

Therefore, we argue that reliable estimation of the summary-worthy content in microblog conversations requires additional consideration of sentiment. Considering that sentiment can also be represented by word distributions and captured via topic models in an unsupervised or weakly-supervised manner~\cite{DBLP:conf/cikm/LinH09,DBLP:conf/wsdm/JoO11,DBLP:conf/cikm/LimB14}. In future work, we can propose another model based on our joint model \textsc{topic+disc+rel} that can additionally separate sentiment word representations from discourse and topics. Besides, \citet{DBLP:conf/acl/LazaridouTS13} has demonstrated that sentiment shifts can indicate sentence-level discourse functions in product reviews. We can then hypothesize that modeling discourse roles of messages can also benefit from exploring sentiment shifts in conversations. As it might be out of the scope of this article to thoroughly explore the joint effects of topic, discourse, and sentiment on microblog conversation summarization, we hence leave the study on such extended model to future work.

%% file: tables/summ_statistics.tex
 \begin{table}[ht]\footnotesize
\begin{center}
\caption{Description of the ten conversation trees for summarization. Each line describes the statistic information of one conversation.}\label{tab:summ-data}
\begin{tabular}{|rrl|}
\hline
\# of messages	&	 Height	&	 Description\\
\hline
21,353	&	16	&	 HKU dropping out student wins the college entrance exam again.\\
9,616	&	11	&	 German boy complains hard schoolwork in Chinese High School.\\	
13,087	&	8	&	 Movie Tiny Times 1.0 wins high grossing in criticism.\\
12,865	&	8	&	 ``I am A Singer'' states that singer G.E.M asking for resinging conforms to rules.\\
10,666	&	8	&	 Crystal Huang clarified the rumor of her derailment.\\	
21,127	&	11	&	 Germany routs Brazil 7:1 in World-Cup semi-final.\\
18,974	&	13	&	 The pretty girl pregnant with a second baby graduated with her master degree. \\
2,021	&	18	&	 Girls appealed for equality between men and women in college admission\\
9,230	&	14	&	 Violent terrorist attack in Kunming railway station.\\
10,052	&	25	&	 MH17 crash killed many top HIV researchers.\\
\hline
\end{tabular}
\end{center}
\end{table}

%% file: tables/rouge_results.tex
\begin{table}
	\begin{center}
		\caption{Average ROUGE for model-produced summaries against the three human-generated references. Len: count of Chinese characters in the extracted summary. Prec, Rec, and F1: average precision, recall, and F1 ROUGE measure over the ten conversation trees (\%). The best scores in each setting is highlighted in \textbf{bold}. Higher scores indicate better results.}	\label{tab:summ}
			\begin{tabular}{|l|l|rrr|rrr|}
				\hline
				\multirow{2}{*}{Models}	&	\multirow{2}{*}{Len}	&
				\multicolumn{3}{c|}{ROUGE-1}& \multicolumn{3}{c|}{ROUGE-2}\\
				\cline{3-8}
				& & Prec&Rec& F1&Prec&Rec& F1\\
				\hline
				\underline{\textbf{Baselines}} &&&&&&&\\
				\textsc{Length}	& 95.4
				&19.6&\textbf{53.2}&28.1
				&5.1&\textbf{14.3}&7.3\\
				\textsc{Popularity}	&27.2
				&33.8&25.3&27.9
				&8.6&6.1&6.8\\
				\textsc{User}	&37.6
				&32.2&34.2&32.5
				&8.0&8.9&8.2\\
				\textsc{LexRank}	&25.7
				&\textbf{35.3}&22.2&25.8	
				&11.7&6.9&8.3\\
				\hline
	\underline{\textbf{State-of-the-art}} &&&&&&&\\
	\textsc{\citet{DBLP:conf/wsdm/ChangWML13}}     &68.6	
				&25.4&48.3&32.8	
				&7.0&13.4&9.1\\
				\textsc{\citet{DBLP:conf/emnlp/LiGWPW15}}	    &58.6
				&27.3&45.4&33.7	
				&7.6&12.6&9.3\\
				\hline
				\underline{\textbf{Our models}} &&&&&&&\\
				\textsc{topic only}	&48.6
				&30.4&40.4&33.6	
				&9.2&12.0&10.0\\
				\textsc{topic+disc}	&37.8
				&\textbf{38.1}&35.5&33.1	
				&\textbf{13.2}&11.5&\textbf{10.8}\\
				\textsc{topic+disc+rel}	&48.9	
				&32.3&41.3&\textbf{34.0}	
				&10.3&12.5&10.5\\
				\hline
				
				\hline
				\hline
				\multirow{2}{*}{Models}	&	\multirow{2}{*}{Len}	&
				\multicolumn{3}{c|}{ROUGE-L}	&	\multicolumn{3}{c|}{ROUGE-SU4}\\
				\cline{3-8}
				& & Prec&Rec& F1&Prec&Rec& F1\\
				\hline
				\underline{\textbf{Baselines}} &&&&&&&\\
				\textsc{Length}	& 95.4
				&16.4&\textbf{44.4}&23.4
				&6.2&\textbf{17.2}&8.9\\
				\textsc{Popularity}	&27.2
				&28.6&21.3&23.6
				&10.4&7.6&8.4\\
				\textsc{User}	&37.6
				&28.0&29.6&28.2	
				&9.8&10.6&10.0\\
				\textsc{LexRank}	&25.7
				&\textbf{30.6}&18.8&22.1	
				&\textbf{12.3}&7.5&8.8\\
				\hline	\underline{\textbf{State-of-the-art}} &&&&&&&\\
				\textsc{\citet{DBLP:conf/wsdm/ChangWML13}} 	    &68.6
				&21.6&41.1&27.9	
				&8.3&16.0&10.8\\
				\textsc{\citet{DBLP:conf/emnlp/LiGWPW15}}	    &58.6                &23.3&38.6&28.7
				&8.8&14.7&10.9\\
				\hline
				\underline{\textbf{Our models}} &&&&&&&\\
				\textsc{topic only}	&48.6
				&26.3&34.9&29.0	
				&10.2&13.8&11.3\\
				\textsc{topic+disc}	&37.8
				&\textbf{33.3}&30.7&28.6	
				&\textbf{13.3}&12.2&11.3\\
				\textsc{topic+disc+rel}	&48.9
				&28.0&35.4&\textbf{29.3}	 &10.9&14.0&\textbf{11.5}\\
				\hline
				
			\end{tabular}
	\end{center}
\end{table}	

%% file: tables/human_results.tex
\begin{table}
	\caption{Overall human ratings on summaries produced by varying summarization systems. Info, Conc, Read are short forms of informativeness, conciseness, and readability, respectively. The ratings are given in a $1$-$5$ Likert scale. Higher scores indicate better ratings. The best score in each setting is highlighted in \textbf{bold}.
	}  \label{tab:rate}
	\begin{center}
		\begin{tabular}{|lrrr|}  
			\hline  
			Models &Info& Conc & Read\\ 
			\hline 
			\underline{\textbf{Baselines}}&&&\\
			\textsc{Length}&2.33&2.93&2.28\\
			\textsc{Popularity}&2.38&2.35&3.05\\
			\textsc{User}&3.13&3.10&3.75\\
			\textsc{LexRank}&3.05&2.70&3.03\\
			\hline
			\hline
			\underline{\textbf{State-of-the-art}}&&&\\
			\textsc{\citet{DBLP:conf/wsdm/ChangWML13}}&3.43&3.50&3.70\\
			\textsc{\citet{DBLP:conf/emnlp/LiGWPW15}}&\textbf{3.70}&\textbf{3.90}&\textbf{4.15}\\
			\hline
			\hline
			\underline{\textbf{Our models}}&&&\\
			\textsc{topic only}&3.33&3.03&3.35\\
			\textsc{topic+disc}&3.25&3.15&3.55\\
			\textsc{topic+disc+rel}&3.35&3.28&3.73\\
			\hline
		\end{tabular}  
	\end{center}
\end{table} 

%% file: tables/summ_case.tex
\begin{CJK}{UTF8}{gbsn}
\begin{table}\small
\begin{center}
\caption{
The sample summary produced by \textsc{topic+disc+rel}-based summarization system. For each message, we display the original Chinese version followed by its English translation.}\label{tab:summary}
\begin{tabular}{|p{12.8cm}|}
\hline
\textbf{\underline{Root message of the conversation:}}\\
近日，各高校招生录取分数纷纷出炉，国际关系学院等院校分数线设置女高男低，引起了广州一位女大学生的关注。她认为这种做法对女考生很不公平，于是写信给国家主席习近平，希望能关注高考录取中的性别歧视现象，重视女性在科技国防军事中的力量。
Recently, the admission criteria for colleges are coming out. Women should get better grades in College Entrance Exam to go to colleges like University of International Relations. A female undergraduate student in Guangzhou was concerned about the unfair treatment. She wrote a letter to President Xi, Jinping for reducing gender discrimination in college admission and emphasized the important role female plays in technology and military.\\
\hline
\hline
\textbf{\underline{The produced summary:}}\\
以保护之名提高女性受教育门槛，实质上是一种“把女性视为弱者”的社会刻板印象作祟，这违背了联合国《消除对妇女一切形式歧视公约》中”保护性歧视“的规定。再者，”本专业需要多熬夜女生吃不消“这一理由并不正当，难道分数线以上考进去的女生的生理健康就不需要保护了吗？分数高的学生更能熬夜？
Raising the bar for women to get education in order to protect them is ascribed to a stereotype of ``\textit{women are weaker sex}''. This is ``\textit{special protections for women}'' in ``\textit{The Convention on the Elimination of all Forms of Discrimination Against Women}'' released by UN. Besides, ``\textit{students in our department should stay up to learn}'' is not an appropriate reason. What about their female students? Don't they have to take care of their physical health? Or students achieving higher grades don't need much sleep?\\
嗯其实…要是没有分数差不限制男女比例的话…学校里男生又会特别少抱怨的还会是我们妹子自己啦…所以…多方面看吧
In fact, we need to use different admission criteria to avoid gender imbalance. If a college has too few boys, girls will complain. Every coin has two sides.\\
因为大学都承认男人就是不如女人啊，呵呵 
Because colleges admit that women are better than men, hehe.\\
以前看到的一则新闻说的是为了调整语言类专业的男女比例，下调男考生的录取分数线。如果像这样以女生体质为借口，那同样寒窗苦读十二载，女生的成绩分量不应该更重才对么？希望习大大能看到吧
An earlier news reported that men could be admitted with worse grades than women for encouraging men to study language. If women do have worse physical condition, then it is more difficult for women to get the same grades as men. Women should have lower bar in college admission. I hope President Xi can see this.\\
说怕体力吃不消严格要求体育分也就罢了，文化分数高低能作为一个人适不适合一个工作辛苦的岗位的理由么？
If they are concerned about physical conditions of women, then they should require a test in PE. Why use paper based exam to test physical conditions?\\
说得好！必须转～
Good point! I have to repost this.\\
哈哈哈，国关课业繁重经常熬夜～
Hahaha, the workload in International Relations is so heavy that students should stay up to learn.\\
女性普遍比男生更努力却换来不同的的结果，要我说男女平等，男性角色弱化无可厚非，抱着几千年的传统观念看今天的男生是不行的，男孩危机是个伪命题，况且真有本事不至于就在学校受益
Generally, women work harder than men but have worse endings. For gender equality, it is alright to weaken the role of men. We should have a different view on the boys today. ``\textit{The boy crisis}'' is nonsense. Besides, you can still be a great guy without education.\\
这难道就是女性为什么越来越优秀，而男性越来越屌丝的部分原因？呵呵。男同胞要感谢性别歧视，让他们越来越弱了。
Isn't this part of the reason why women become more and more excellent while men go to the opposite direction? Interesting. Men should appreciate for sexism, which makes them weaker and weaker.\\
\hline
\end{tabular}
\end{center}
\end{table}
\end{CJK}

%% file: sections/conclusion.tex
\section{Conclusion and Future Work}

In this article, we have presented a novel topic model for microblog messages that allows the joint induction of conversational discourse and latent topics in a fully unsupervised manner.
By comparing our joint model with a number of competitive topic models on real-world microblog datasets, we have demonstrated the effectiveness of using conversational discourse structure to help in identifying topical content embedded in short and colloquial microblog messages.
Moreover, our empirical study on microblog conversation summarization has shown that the produced discourse and topical representations can also predict summary-worthy content.
Both ROUGE evaluation and human assessment have demonstrated that the summaries generated based on the outputs of our joint model are informative, concise, and easy-to-read.
Error analysis on the produced summaries has shown that sentiment should be effectively captured and separated to further advance our current summarization system forward. 
As a result, the joint effects of discourse, topic, and sentiment on microblog conversation summarization is worthy exploring in future study.

For other lines of future work, one potential is to extend our joint model to identify topic hierarchies from microblog conversation trees.
In doing so, one could learn how topics change in a microblog conversation along with a certain hierarchical path.
Another potential line is to combine our work with representation learning on social media.
Although some previous studies have provided intriguing approaches to learning  representations at the level of words~\cite{DBLP:conf/nips/MikolovSCCD13,DBLP:conf/naacl/MikolovYZ13}, sentences~\cite{DBLP:conf/icml/LeM14}, and paragraphs~\cite{DBLP:conf/nips/KirosZSZUTF15}, they are limited in modeling social media content with colloquial relations.
Following similar ideas in this work, where discourse and topics are jointly explored, we can conduct other types of representation learning, e.g., embeddings for words~\cite{DBLP:conf/icwsm/LiSLN17}, messages~\cite{DBLP:conf/acl/DhingraZFMC16}, or users~\cite{DBLP:conf/emnlp/DingBP17}, in context of conversations, which should complement social media representation learning and vice versa.

%% file: sections/appendix.tex
%\appendix
\section*{Appendix}

\input{sections/inference.tex}

\input{tables/notations_inference.tex}

%% file: sections/inference.tex
In this section, we present the key steps for inferring our joint model of conversational discourse and latent topics. Its generation process has been described in Section~\ref{sec:model}.
As described in Section~\ref{sec:model}, we employ collapsed Gibbs sampling~\cite{DBLP:conf/nips/GriffithsSBT04} for model inference. Before giving formula of sampling steps, we first define the notations of all variables used in the formulations of Gibbs sampling, which are described in Table \ref{tab:notions}. In particular, the various $\mathbb{C}$ variables refer to counts excluding the message $m$ on conversation tree $c$.

For each message $m$ on conversation tree $c$, we sample its discourse role $d_{c,m}$ and topic assignment $z_{c,m}$ according to the following conditional probability distribution:

\begin{equation}\small
	\begin{aligned}
		&	p(d_{c,m}=d, z_{c,m}=k|\mathbf{d}_{\neg{(c,m)}},\mathbf{z}_{\neg{(c,m)}}\mathbf{w},\mathbf{x},\Theta)\\
		\propto &
\frac{\Gamma(\mathbb{C}^{DD}_{d_{c,pa(m)},(\cdot)}+D\cdot\gamma)}{\Gamma(\mathbb{C}^{DD}_{d_{c,pa(m)},(\cdot)}+I(d_{c,pa(m)}\neq d)+D\cdot\gamma)} %\\
		\cdot	%&
		\frac{\Gamma(\mathbb{C}^{DD}_{d_{c,pa(m)},(d)}+I(d_{c,pa(m)}\neq d)+\gamma)}{\Gamma(\mathbb{C}^{DD}_{d_{c,pa(m)},(d)}+\gamma)} \\
		\cdot	& \frac{\Gamma(\mathbb{C}^{DD}_{d,(\cdot)}+D\cdot\gamma)}{\Gamma(\mathbb{C}^{DD}_{d,(\cdot)}+I(d_{c,pa(m)}=d)+\mathbb{N}^{DD}_{(\cdot)}+D\cdot\gamma)} %\\
		\cdot	%& 
\prod_{d'=1}^D\frac{\Gamma(\mathbb{C}^{DD}_{d,(d')}+\mathbb{N}^{DD}_{(d')}+I(d_{c,pa(m)}=d=d')+\gamma)}{\Gamma(\mathbb{C}^{DD}_{d,(d')}+\gamma)}\\
\cdot &
\frac{\mathbb{C}^{CT}_{c,(k)}+\alpha}{\mathbb{C}^{CT}_{c,(\cdot)}+K\cdot\alpha}
\cdot\frac{\Gamma(\mathbb{C}^{TW}_{k,(\cdot)}+V\cdot\beta)}{\Gamma(\mathbb{C}^{TW}_{k,(\cdot)}+\mathbb{N}^{TW}_{(\cdot)}+V\cdot\beta)}\cdot\prod_{v=1}^{V}\frac{\Gamma(\mathbb{C}^{TW}_{k,(v)}+\mathbb{N}^{TW}_{(v)}+\beta)}{\Gamma(\mathbb{C}^{TW}_{k,(v)}+\beta)}
\\
\cdot &
\frac{\Gamma(\mathbb{C}^{DX}_{d,(\cdot)}+3\cdot\delta}{\Gamma(\mathbb{C}^{DX}_{d,(\cdot)}+\mathbb{N}^{DX}_{(\cdot)}+3\cdot\delta)}\cdot\prod_{x=1}^3\frac{\Gamma(\mathbb{C}^{DX}_{d,(x)}+\mathbb{N}_{(x)}^{DX}+\delta)}{\Gamma(\mathbb{C}^{DX}_{d,(x)}+\delta)}\\
		\cdot	&	\frac{\Gamma(\mathbb{C}^{DW}_{d,(\cdot)}+V\cdot\beta)}{\Gamma(\mathbb{C}^{DW}_{d,(\cdot)}+\mathbb{N}^{DW}_{(\cdot)}+V\cdot\beta)}\cdot\prod_{v=1}^{V}\frac{\Gamma(\mathbb{C}^{DW}_{d,(v)}+\mathbb{N}^{DW}_{(v)}+\beta)}{\Gamma(\mathbb{C}^{DW}_{d,(v)}+\beta)} \\
\end{aligned}\label{eq:message-sample}
\end{equation}
where the discourse role and topic assignments of message $m$ on conversation $c$ are determined by: 1) the discourse role assignments of the parent and all the children of message $m$ on conversation $c$ (shown in the first $4$ factors); 2) the topic mixture of conversation tree $c$ (shown in the $5$-th factor); 3) the topic assignments of other messages sharing TOPIC words with $m$ (shown in the $6$-th and the $7$-th factor); 4) the distribution of words in $m$ as DISC, TOPIC, and BACK words (shown in the $8$-th and the $9$-th factor); 5) the discourse role assignments of other messages sharing DISC words with $m$ (shown in the last two factors).

For each word $n$ in $m$ on $c$, the sampling formula of its word type $x_{c,m,n}$ (as discourse (DISC), topic (TOPIC), and background (BACK)) is given as the following:

\begin{equation}\small
	\begin{aligned} \label{eq:word-sample}
		&	p(x_{c,m,n}=x|\mathbf{x}_{\neg{(c,m,n)}},\mathbf{d},\mathbf{z},\mathbf{w},\Theta) \\
		\propto		&	\frac{\mathbb{C}^{DX}_{d_{c,m},(x)}+\delta}{\mathbb{C}^{DX}_{d_{c,m},(\cdot)}+3\cdot\delta}\cdot g(x,c,m,n)\\
	\end{aligned} 
\end{equation}
where
{\small
\begin{eqnarray}
		g(x,c,m,n)=
		\begin{cases}
\frac{\mathbb{C}^{DW}_{d_{c,m},(w_{c,m,n})}+\beta}{\mathbb{C}^{DW}_{d_{c,m},(\cdot)}+V\cdot\beta}	& \text{if }x==\text{DISC}\\			\frac{\mathbb{C}^{TW}_{z_{c,m},(w_{c,m,n})}+\beta}{\mathbb{C}^{TW}_{z_{c,m},(\cdot)}+V\cdot\beta}	&	\text{if } x==\text{TOPIC}\\ 			\frac{\mathbb{C}^{BW}_{(w_{c,m,n})}+\beta}{\mathbb{C}^{BW}_{(\cdot)}+V\cdot\beta}	& \text{if }x==\text{BACK}\\ 
		\end{cases}
	\end{eqnarray}
}

\noindent Here the word type switcher of word $n$ in message $m$ on conversation $c$ is determined by: 1) The distribution of word types in messages sharing the same discourse role as $m$ (shown in the first factor); 2) And the word types of word $w_{c,m,n}$ appearing elsewhere (shown in the second factor $g(x,c,m,n)$).

%% file: tables/notations_inference.tex
  \begin{table}\small
    \begin{center}
        \caption{The notations of symbols in the sampling formulas Eq. \ref{eq:message-sample} and Eq. \ref{eq:word-sample}. $(c,m)$: message $m$ on conversation tree $c$.}\label{tab:notions}
    \begin{tabular}{|c|m{11.5cm}|}
 		\hline
 		$x$ &   word-level word type switcher. $x=1$: discourse word (DISC); $x=2$: topic word (TOPIC); $x=3$: background word (BACK).\\
		\hline
		$\mathbb{C}^{DX}_{d,(x)}$ 	&	\# of words with word type as $x$ and occurring in messages with discourse $d$.\\
		\hline
		$\mathbb{C}^{DX}_{d,(\cdot)}$ &	\# of words that occur in messages whose discourse assignments are $d$, i.e., $\sum_{x=1}^3 \mathbb{C}^{DX}_{d,(x)}$. \\
		\hline
		$\mathbb{N}_{(x)}^{DX}$	&	\# of words occurring in message $(c,m)$ and with word type assignment as $x$. \\
		\hline
		$\mathbb{N}_{(\cdot)}^{DX}$		&	\# of words in message $(c,m)$, i.e., 
		$\mathbb{N}_{(\cdot)}^{DX}=\sum_{x=1}^3 \mathbb{N}_{(x)}^{DX}$. \\
		\hline
		$\mathbb{C}^{DW}_{d,(v)}$	&	\# of words indexing $v$ in vocabulary, assigned as discourse word, and occurring in messages assigned discourse $d$.\\
		\hline
		$\mathbb{C}^{DW}_{d,(\cdot)}$	&	\# of words assigned as discourse words (DISC) and occurring in messages assigned as discourse $d$, i.e., $\mathbb{C}^{DW}_{d,(\cdot)}=\sum_{v=1}^V \mathbb{C}^{DW}_{d,(v)}$.\\
		\hline
		$\mathbb{N}_{(v)}^{TW}$	&	\# of words indexing $v$ in vocabulary that occur in messages $(c,m)$ and are assigned as topic words (TOPIC). \\
		\hline
		$\mathbb{N}_{(\cdot)}^{TW}$	&	\# of words assigned as topic words (TOPIC) and occurring in message $(c,m)$, i.e., $\mathbb{N}_{(\cdot)}^{TW}=\sum_{v=1}^V \mathbb{N}_{(v)}^{TW}$. \\
		\hline
        $\mathbb{N}_{(v)}^{DW}$	&	\# of words indexing $v$ in vocabulary that occur in messages $(c,m)$ and are assigned as discourse words (DISC). \\
		\hline
		$\mathbb{N}_{(\cdot)}^{DW}$	&	\# of words assigned as discourse words (DISC) and occurring in message $(c,m)$, i.e., $\mathbb{N}_{(\cdot)}^{DW}=\sum_{v=1}^V \mathbb{N}_{(v)}^{DW}$. \\
		\hline
		$\mathbb{C}^{DD}_{d,(d')}$	&	\# of messages assigned discourse $d'$ whose parent is assigned discourse $d$.\\
		\hline
		$\mathbb{C}^{DD}_{d,(\cdot)}$	&	\# of messages whose parents are assigned discourse $d$, i.e., $\mathbb{C}^{DD}_{d,(\cdot)}=\sum_{d'=1}^{D}\mathbb{C}^{DD}_{d,(d')}$.\\
		\hline
		$I(\cdot)$	&	An indicator function, whose value is 1 when its argument inside $()$ is true, and 0 otherwise. \\
		\hline
		$\mathbb{N}^{DD}_{(d)}$	&	\# of messages whose parent is $(c,m)$ and assigned discourse $d$.\\
		\hline
		$\mathbb{N}^{DD}_{(\cdot)}$	&	\# of messages whose parent is $(c,m)$, i.e.,  $\mathbb{N}^{DD}_{(\cdot)}=\sum_{d=1}^{D}\mathbb{N}^{DD}_{(d)}$\\
		\hline
		$\mathbb{C}^{BW}_{(v)}$	&	\# of words indexing $v$ in vocabulary and assigned as background words (BACK)\\
		\hline
		$\mathbb{C}^{BW}_{(\cdot)}$	&	\# of words assigned as background words (BACK), i.e., $\mathbb{C}^{BW}_{(\cdot)}=\sum_{v=1}^{V}\mathbb{C}^{BW}_{(v)}$\\
		\hline
        $\mathbb{C}^{CT}_{c,(k)}$	&	\# of messages on conversation tree $c$ and assigned topic $k$.\\
\hline
$\mathbb{C}^{CT}_{c,(\cdot)}$	&	\# of messages on conversation tree $c$, i.e., $\mathbb{C}^{CT}_{c,(\cdot)}=\sum_{k=1}^{K}\mathbb{C}^{CT}_{c,(k)}$\\
\hline 
$\mathbb{C}^{TW}_{k,(v)}$	&	\# of words indexing $v$ in vocabulary, sampled as topic words (TOPIC), and occurring in messages assigned topic $k$.\\
\hline
$\mathbb{C}^{TW}_{k,(\cdot)}$	&	\# of words assigned as topic word and occurring in messages assigned topics $k$ (TOPIC), i.e., $\mathbb{C}^{TW}_{k,(\cdot)}=\sum_{v=1}^V \mathbb{C}^{TW}_{k,(v)}$.\\
\hline
	\end{tabular}
    \end{center}
    \end{table}

%% file: sections/ack.tex
\begin{acknowledgments}
This work is partially supported by Innovation and Technology Fund (ITF) Project No. 6904333, General Research Fund (GRF) Project No. 14232816 (12183516), National Natural Science Foundation of China (Grant No. 61702106), and Shanghai Science and Technology Commission (Grant No. 17JC1420200 and Grant No. 17YF1427600). 
We are grateful for the contributions of Yulan He, Lu Wang, and Wei Gao in shaping part of our ideas, and the efforts of Nicholas Beautramp, Sarah Shugars, Ming Liao, Xingshan Zeng, Shichao Dong, and Dingmin Wang in preparing some of the our experiment data.
Also, we thank Shuming Shi, Dong Yu, Tong Zhang, and the three anonymous reviewers for the insightful suggestions on various aspects of this work.
\end{acknowledgments}